\documentclass{article}

\IfFileExists{arxiv.sty}{%
  \usepackage{arxiv}%
}{%
  \providecommand{\undertitle}{}%
  \providecommand{\headeright}{}%
}
\usepackage[utf8]{inputenc}
\usepackage[T1]{fontenc}
\usepackage[authoryear,round]{natbib}
\usepackage{hyperref}
\usepackage{url}
\usepackage{booktabs}
\usepackage{amsmath}
\usepackage{amssymb}
\usepackage{microtype}
\usepackage[table]{xcolor}
\usepackage{graphicx}
\usepackage{fontawesome5}
\usepackage{enumitem}
\usepackage{multirow}
\usepackage{array}
\usepackage{tabularx}
\usepackage{tcolorbox}
\tcbuselibrary{breakable}
\usepackage{caption}
\usepackage{float}
\usepackage{pifont}
\usepackage{xcolor}
\usepackage{letltxmacro}

\newcommand{\cmark}{\textcolor{black}{\ding{51}}}
\newcommand{\xmark}{\textcolor{gray!70}{\ding{55}}}

\definecolor{factblue}{RGB}{53,104,165}
\definecolor{sectionblue}{RGB}{68,132,190}
\definecolor{skillgreen}{RGB}{48,135,99}
\definecolor{structurepurple}{RGB}{115,77,145}
\definecolor{lightgray}{RGB}{247,248,250}
\definecolor{tableheader}{RGB}{222,220,232}
\definecolor{citationgreen}{RGB}{0,110,51}

\renewenvironment{abstract}{%
  \begin{tcolorbox}[
    breakable,
    colback=factblue!4!white,
    colframe=factblue!35!white,
    boxrule=0.5pt,
    arc=2mm,
    left=12pt,right=12pt,top=7pt,bottom=3pt,
    before skip=14pt,after skip=16pt
  ]%
  {\centering\large\bfseries\upshape\color{black}Abstract\par}%
  \vspace{8pt}\linespread{1.06}\selectfont\noindent\ignorespaces
}{%
  \end{tcolorbox}%
}

\hypersetup{
  colorlinks=true,
  linkcolor=structurepurple,
  citecolor=blue,
  urlcolor=factblue
}

\setcitestyle{aysep={,},notesep={, }}
\LetLtxMacro\cite\citep
\DeclareRobustCommand{\referencehighlight}[1]{%
  \leavevmode
  \begingroup
  \hypersetup{linkcolor=red}%
  \color{black}#1%
  \endgroup
}

\newcolumntype{Y}{>{\raggedright\arraybackslash}X}
\newcolumntype{L}[1]{>{\raggedright\arraybackslash}p{#1}}

\renewcommand{\undertitle}{}
\renewcommand{\headeright}{}
\providecommand{\shorttitle}{}
\renewcommand{\shorttitle}{\normalfont\footnotesize A Benchmark for Evaluating Self-Evolving Agents in Long-Horizon Interdependent Task Streams}
\date{}

\makeatletter
\newcommand{\numberedauthorbibliography}{%
  \let\@biblabel\NAT@biblabelnum
  \let\@bibsetup\NAT@bibsetnum
}
\renewcommand{\section}{%
  \@startsection{section}{1}{\z@}%
    {-2.0ex \@plus -0.5ex \@minus -0.2ex}%
    {1.5ex \@plus 0.3ex \@minus 0.2ex}%
    {\large\bfseries\raggedright\color{sectionblue}}%
}
\@ifundefined{@toptitlebar}{}{%
  \let\originaltoptitlebar\@toptitlebar
  \renewcommand{\@toptitlebar}{%
    \originaltoptitlebar
    \vbox to 0pt{\vskip 35mm
      \hbox to \textwidth{%
        \hskip 14.5cm
        \includegraphics[height=1.6cm]{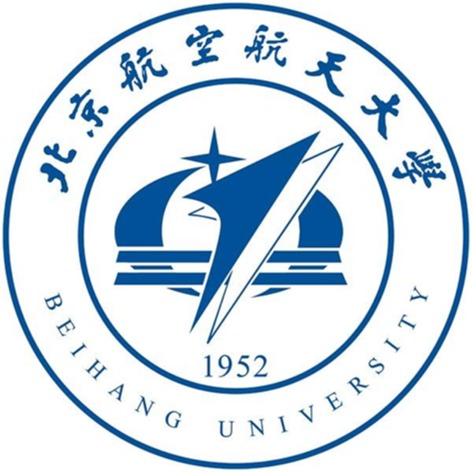}\hfill}%
      \vss
    }%
  }%
}
\@ifundefined{@bottomtitlebar}{}{%
  \let\originalbottomtitlebar\@bottomtitlebar
  \renewcommand{\@bottomtitlebar}{\originalbottomtitlebar\vskip -18pt}%
}
\makeatother

\title{EvoLink-Bench: A Benchmark for Evaluating Self-Evolving \\ Agents in Long-Horizon Interdependent Task Streams}

\author{%
  {\normalfont Wei Chen\textsuperscript{\dag}} \quad {\normalfont Xiaodong Lu\textsuperscript{\dag}} \quad {\normalfont Weihao Shen\textsuperscript{\dag}}  \quad
  {\normalfont Gongxun Li\textsuperscript{\dag}} \quad {\normalfont Xingyu Guo\textsuperscript{\dag}} \quad \normalfont Qihang Sun\textsuperscript{\dag} \\ \\[-0.8em] Fuwei Zhang \quad 
  Zhao Zhang   \quad Yikun Ban \quad Shuang Li  \quad 
  Deqing Wang \quad Fuzhen Zhuang \\[0.5em]
  \faIcon{university}\enspace Beihang University\quad {\small\mdseries\faIcon{calendar-alt}\enspace\textbf{Date:} November 24, 2026} \\ [0.5em]
  {\small\mdseries\color{black}\hypersetup{urlcolor=blue}%
    \begin{tabular}{@{}l@{}}
      \makebox[1.3em][c]{\faGithub}\enspace \textbf{Code Repository:} {\mdseries\urlstyle{same}\url{https://github.com/BUAA-Self-Evolving/EvoLink-Bench}} \\[0.25em]
      \makebox[1.3em][c]{\faEnvelope}\enspace\textbf{Contact:} \textcolor{black}{\mdseries\{chenwei23,xiaodonglu,22373140\}@buaa.edu.cn}
    \end{tabular}%
  }
}

\begin{document}

\setlength{\headheight}{23pt}
\addtolength{\topmargin}{-9pt}

\vspace*{-26pt}
\maketitle
\begingroup
\renewcommand{\thefootnote}{\dag}
\footnotetext[2]{These authors contributed equally to this work.}
\endgroup

\vspace{-20pt}
\noindent\includegraphics[width=\linewidth]{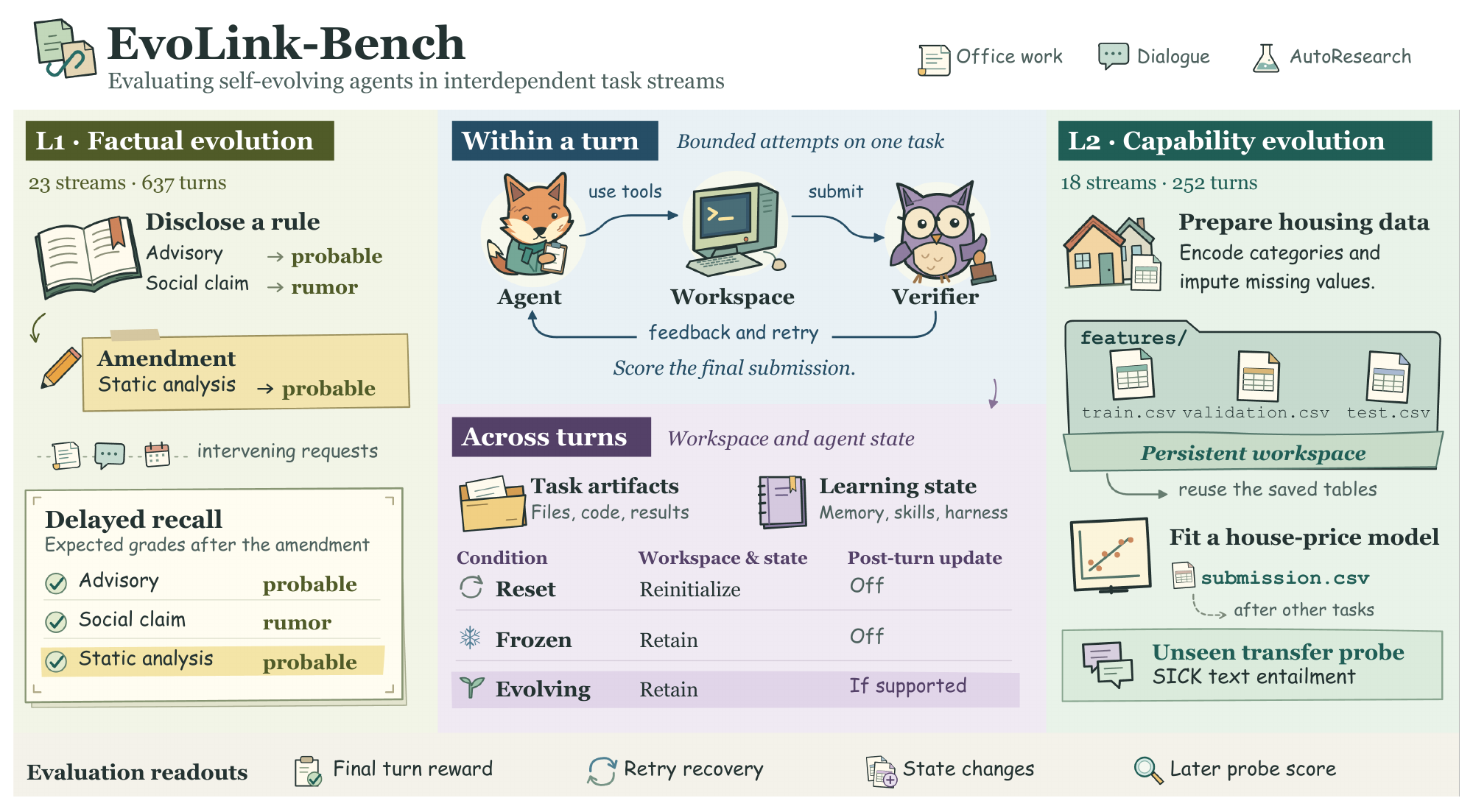}\par
\begingroup
\captionsetup{type=figure,skip=5pt}
\captionof{figure}{Overview of EvoLink-Bench: L1 factual and L2 capability evolution in interdependent task streams. Evaluation separates retries from persistent updates and measures task scores, recovery, state changes, and reuse.}
\label{fig:evolink-overview}
\endgroup

\begin{abstract}
	Large language model (LLM) agents are evolving from passive chatbots into autonomous systems that operate continuously in dynamic environments. Such agents need not only solve current tasks but also learn from past interactions to improve future performance. However, existing streaming benchmarks often rely on short task sequences or largely independent tasks, providing limited evidence of whether agents can effectively reuse experience when later tasks depend on earlier outcomes.
	To address this gap, we introduce \textbf{EvoLink-Bench}, a benchmark for evaluating self-evolving agents in long-horizon, interdependent task streams. It covers daily office work, multi-turn dialogue, and AutoResearch-style workflows, with persistent workspaces and dependencies across tasks. The benchmark includes two complementary levels: \textbf{L1}, which evaluates the acquisition and revision of language-expressible knowledge, and \textbf{L2}, which assesses the development of task-level capabilities through execution and feedback. Its evaluation further distinguishes within-task recovery, persistent-state updates, and experience reuse in later tasks.
	We evaluate five agent frameworks across three backbone configurations under three persistence conditions, covering 19,635 scored task slots from selected evaluation cohorts. Results show that no framework consistently performs best across both levels, retries recover only a limited proportion of failed tasks, and persistent-state updates do not necessarily improve subsequent performance. Furthermore, the effects of evolution differ between knowledge- and capability-level tasks, suggesting that these two forms of evolution require different learning mechanisms. Performance gains also vary across domains and task streams, with improvements on some tasks not necessarily extending to others. These findings highlight the need to evaluate not only whether agents accumulate experience, but also whether they can effectively apply it to future tasks.
\end{abstract}

\section{Introduction}
\label{sec:introduction}

Large language model (LLM) agents are rapidly evolving from stateless problem solvers into interactive, lifelong assistants \cite{wang2024survey, xi2023rise}. In real-world deployments, agents are expected to do more than answer isolated queries; they need to assist users continuously over extended periods. During this process, interactions can generate valuable experience, such as user preferences revealed through dialogue, corrections to tool usage, or newly created task artifacts. This calls for self-evolving agents that can not only complete immediate tasks but also learn from past interactions to improve future decision-making \cite{wang2023voyager, majumder2024clin}. The ability to learn continuously during service is therefore a key requirement for practical self-evolving agents.

Such continuous deployment creates a complex and highly interconnected environment. Unlike a sequence of independent tasks, the user's history, the external environment, and the agent's workspace may all change over time. For example, an initial data analysis task may produce a table that serves as the basis for a subsequent strategic decision, while an automated research workflow may generate code and checkpoints that inform a later experimental hypothesis \cite{lu2024ai}. Moreover, later tasks may challenge earlier assumptions, require the agent to revisit a capability after a long interval, or introduce unrelated tasks that compete for its attention. To operate effectively in such environments, agents must not only retain and reuse relevant experience but also revise outdated knowledge and manage interference, while maintaining the consistency of the persistent workspace on which future tasks depend.

As research on self-evolving agents continues to advance, evaluating their capabilities in realistic continual-service settings has become increasingly important. Existing benchmarks cover individual agent capabilities and application-specific settings~\cite{yehudai2026evaluation}, but address only parts of this broader problem. Short streaming benchmarks typically use repeated tasks to evaluate local transfer and reflection \cite{jiang2026seaeval}. Longer task streams can reveal experience accumulation and cross-domain interference, but are often constructed by concatenating independently specified tasks, without requiring later tasks to depend on earlier outputs or decisions \cite{wei2025evomemory, yan2026agentstream}. Meanwhile, some stateful interactive benchmarks preserve tools and world states, but focus on narrower scenarios with task-specific protocols \cite{trivedi2024appworld, yao2024taubench, cai2025ell, fu2026traineebench}. As a result, existing benchmarks provide limited insight into whether agents can continuously improve through experience across diverse, long-running interactions, particularly when later tasks depend on earlier decisions and artifacts.

To address this gap, we introduce \textbf{EvoLink-Bench}, a benchmark designed to evaluate self-evolving agents in long-horizon, interdependent task streams. Figure~\ref{fig:evolink-overview} illustrates its overall design, including two evolution levels, a persistent task environment, and a unified evaluation protocol. Unlike benchmarks that evaluate isolated tasks or simply concatenate independent examples, EvoLink-Bench organizes tasks into persistent streams, where user context and workspace artifacts carry across interactions. It covers daily office work, multi-turn dialogue, software and data tasks, and automated research workflows. More importantly, tasks are connected through explicit dependencies: later tasks may require agents to recall earlier facts and decisions or reuse previously generated code, data, and experimental results. By varying the intervals between related tasks and introducing intervening tasks, the benchmark examines whether agents can reuse and transfer experience, revise outdated knowledge, and handle interference over time.

EvoLink-Bench evaluates self-evolution at two complementary levels. \textbf{L1} focuses on knowledge evolution, examining whether agents can acquire, retain, apply, and revise information that can be expressed in language, such as facts, preferences, and procedural rules. \textbf{L2} focuses on capability evolution, examining whether agents can develop reusable task-solving procedures through execution and feedback, including coding, data analysis, and research workflows. This distinction captures two different aspects of learning: knowing what to remember and improving how to perform tasks. To evaluate these processes, we introduce three persistence conditions: Reset, Frozen, and Evolving, which differ in whether previous state is retained and whether post-turn updates are enabled. With bounded attempts, our evaluation further separates within-task recovery from changes to persistent agent state and their potential benefits on later probe tasks. This allows us to examine not only whether agents update their state, but also whether such updates are associated with improved performance on subsequent tasks.

We evaluate five agent frameworks across three backbone configurations under these persistence conditions. The results reveal that self-evolution is not a uniform source of performance gains: its effects depend on the agent framework, the level of evolution, and the task distribution. In particular, knowledge evolution and capability evolution exhibit different performance patterns, suggesting that improvements in retaining information do not necessarily translate into better task-solving capabilities. Moreover, recovering from failure through retries is different from learning across tasks, and changes to persistent state do not necessarily lead to effective experience reuse. These findings expose an important gap between accumulating experience and turning it into lasting improvements, highlighting the need for more comprehensive evaluation of self-evolving agents.
Our main contributions are summarized as follows:
\begin{enumerate}[leftmargin=*]
	\item[\ding{182}] \textbf{\textit{Evaluation Paradigm.}} We establish an evaluation paradigm for self-evolving agents in continuous-service settings, distinguishing within-task recovery, persistent-state updates, and cross-task experience reuse. This provides a systematic way to examine whether improvements extend beyond the current task.
	
	\item[\ding{183}] \textbf{\textit{Two-Level Benchmark.}} We introduce EvoLink-Bench, featuring two complementary levels of evolution: L1 for knowledge evolution and L2 for capability evolution. Its long-horizon task streams incorporate explicit task dependencies, persistent artifacts, varying time intervals, and interference, covering diverse real-world workflows.
	
	\item[\ding{184}] \textbf{\textit{Empirical Findings.}} Through a comprehensive evaluation of agent frameworks under different persistence conditions, we show that self-evolution benefits are highly dependent on the agent and task setting. In particular, persistent-state updates do not necessarily lead to effective downstream reuse, and knowledge-level improvements do not guarantee capability-level gains.
\end{enumerate}

\section{Related Work}
\begin{table*}[!htbp]
	\centering
	\caption{
		Comparison of benchmark design coverage.
		\cmark{} indicates an explicitly instantiated task requirement
		or evaluation condition; \xmark{} indicates that the reviewed
		protocol does not instantiate that criterion.
		Cross-task dependencies require later tasks to use earlier task
		outputs or decisions, beyond sharing skills.
		Controlled scheduling denotes explicit variation of task order,
		spacing, or interference.
		Coverage is assessed at the benchmark-suite level.
	}
	\label{tab:benchmark-comparison}
	\large
	\setlength{\tabcolsep}{9pt}
	\renewcommand{\arraystretch}{1.1}
	\resizebox{\textwidth}{!}{%
		\begin{tabular}{lcccc}
			\toprule
			\rowcolor{tableheader}
			\multicolumn{1}{c}{\raisebox{0.5\normalbaselineskip}[0pt][0pt]{\textbf{Benchmark}}}
			& \shortstack{\textbf{Cross-task}\\
				\textbf{Dependencies}}
			& \shortstack{\textbf{Controlled Stream}\\
				\textbf{Scheduling}}
			& \shortstack{\textbf{Knowledge}\\
				\textbf{Revision}}
			& \shortstack{\textbf{Procedural}\\
				\textbf{Experience Reuse}} \\
			\midrule[\heavyrulewidth]
			AgentBench~\cite{liu2023agentbench}
			& \xmark & \xmark & \xmark & \xmark \\
			StreamBench~\cite{wu2024streambench}
			& \xmark & \cmark & \xmark & \cmark \\
			LifelongAgentBench~\cite{zheng2025lifelongagentbench}
			& \xmark & \xmark & \xmark & \cmark \\
			SEA-Eval~\cite{jiang2026seaeval}
			& \xmark & \xmark & \xmark & \cmark \\
			ClawArena~\cite{ji2026clawarena}
			& \cmark & \xmark & \cmark & \xmark \\
			EvoArena~\cite{xu2026evoarena}
			& \cmark & \xmark & \cmark & \cmark \\
			AgentStream~\cite{yan2026agentstream}
			& \xmark & \cmark & \xmark & \cmark \\
			PAST-Bench~\cite{xue2026pastbench}
			& \cmark & \xmark & \cmark & \cmark \\
			FinEvo-Bench~\cite{deng2026finevobench}
			& \xmark & \cmark & \xmark & \cmark \\
			EvoPathBench~\cite{lin2026evopathbench}
			& \xmark & \cmark & \cmark & \cmark \\
			ScienceClaw-Eval~\cite{zhang2026scienceclaw}
			& \xmark & \xmark & \xmark & \cmark \\
			\midrule
			\textbf{EvoLink-Bench (Ours)}
			& \cmark & \cmark & \cmark & \cmark \\
			\bottomrule
		\end{tabular}%
	}
\end{table*}
\subsection{Continual Learning}
\label{sec:related-continual-learning}

Continual learning investigates how systems assimilate sequential experiences while preserving previously acquired knowledge and transferring it to novel problems \cite{wang2024continualsurvey}. The central challenge lies in the stability-plasticity dilemma: integrating new capabilities under limited resources without catastrophically forgetting older ones. Representative approaches typically address this by either protecting critical parameters that support earlier tasks, such as in elastic weight consolidation \cite{kirkpatrick2017overcoming}, or by actively revisiting stored experiences during subsequent learning phases \cite{rolnick2019experience}. These mechanisms tackle interference from different angles, establishing retention, adaptation, and transfer as complementary criteria for assessing a lifelong learner.
The evaluation protocols in continual learning largely dictate which learning dynamics are exposed. Traditional settings, such as task-, domain-, and class-incremental learning, vary the availability of task identity and the specific boundaries a model must recognize \cite{vandeven2022three}. Furthermore, continual learning extends into reinforcement learning, where task boundaries are absent and experience replay supports learning from ongoing interactions \cite{rolnick2019experience}. 

For large language models (LLMs), continual learning spans across pretraining, instruction tuning, and alignment. Given the prohibitive costs of repeated training, there is a strong motivation for efficient adaptation strategies \cite{wu2024continualsurvey}. At the agent level, learning can also manifest through persistent textual states. For instance, CLIN updates a memory of causal abstractions across trials without modifying the underlying model parameters \cite{majumder2024clin}. Building upon these perspectives, EvoLink-Bench evaluates both explicit knowledge revision and reusable task-solving behaviors within a persistent workspace. Its protocol rigorously disentangles within-task corrections, retained execution states, and post-turn learning updates, allowing the specific outcomes of each mechanism to be examined independently.

\subsection{Self-Evolving Agents}

Self-evolving agents augment the traditional perception-reasoning-action loop by integrating persistent updates derived from ongoing interactions. This experience-driven lifelong learning process is typically organized around exploration, long-term memory management, skill acquisition, and knowledge internalization \cite{cai2025ell}. Existing methodologies primarily diverge in what specific information they choose to retain and revise.

An adjacent line of work improves the current output through inference-time refinement. Self-Refine alternates feedback generation and revision using the same LLM, without additional model training \cite{madaan2023selfrefine}. This establishes output refinement as a distinct evaluation target: improving the current answer does not by itself demonstrate transfer to a subsequent task. EvoLink-Bench therefore separates within-turn retries, post-turn updates, and later probes.

Memory-oriented methods focus on preserving observations, trajectories, and failure lessons to inform future decisions. For example, Reflexion converts task feedback into verbal reflections stored in an episodic memory for subsequent trials \cite{shinn2023reflexion}. Similarly, ExpRAG retrieves past task experiences, while ReMem integrates reasoning, task execution, and memory refinement into a cohesive action-think-memory cycle \cite{wei2025evomemory}. These operations enable the agent to maintain a continuous thread of experience, linking newly observed feedback directly to immediate reasoning and future memory retrieval.
Skill-oriented methods, conversely, abstract experience into reusable procedural knowledge or executable routines. Voyager maintains a dynamically expanding library of code-based skills, improving candidate programs through environmental feedback, execution errors, and self-verification \cite{wang2023voyager}. Once validated, these programs serve as foundational building blocks for future tasks. The ELL framework similarly distills recurring patterns into skills that are refined for novel scenarios \cite{cai2025ell}. AgentSkillOS further addresses the organization of these libraries by utilizing capability-tree retrieval and composing skills into directed acyclic workflows \cite{li2026agentskillos}. 
Another prominent line of research focuses on evolving the agent's harness, which includes prompts, contextual instructions, toolsets, control logic, and verification procedures. ACE treats the agent's context as an evolving playbook, using generation, reflection, and curation to accumulate strategies through structured updates \cite{zhang2025ace}. HarnessBank pairs a task-execution agent with a separate "evolver" that diagnoses failures and systematically generates, recombines, and screens candidate harnesses \cite{luo2026harnessbank}. At this level, experience fundamentally alters how the agent plans and invokes tools across interactions.

Despite these significant advances, existing evaluations fail to fully capture the complexity of self-evolution in truly persistent environments, where later tasks may depend on earlier decisions and artifacts. To bridge this gap and advance research in this setting, we introduce EvoLink-Bench, a benchmark specifically designed to evaluate self-evolving agents within long-horizon, interdependent task streams.

\subsection{Self-Evolving Agent Benchmarks}
Evaluating self-evolving agents requires measuring an agent's trajectory of improvement rather than its static performance. This introduces a fundamental challenge in benchmark design: how to rigorously assess the accumulation, adaptation, and transfer of knowledge without confounding these effects with the agent's baseline capabilities. To address this, existing benchmarks must navigate an inherent trade-off between isolating specific learning mechanisms and simulating realistic, long-horizon environments. Consequently, recent evaluation frameworks have approached this challenge by prioritizing different dimensions of the learning process.

Long-term conversational memory provides a complementary evaluation perspective. LongMemEval distinguishes information extraction, multi-session reasoning, temporal reasoning, knowledge updates, and abstention \cite{wu2025longmemeval}. These abilities inform the diagnostic scope of L1, while our L2 setting additionally evaluates execution and the reuse of task artifacts. Separately, agent-evaluation research emphasizes cost control, adequate holdout tasks, and reproducible evaluation practices \cite{kapoor2024agentsmatter}; we therefore report cohort scope and budgets.

First released in 2023, AgentBench evaluates multi-turn task execution across eight interactive environments \cite{liu2023agentbench}. It provides a reference for task-solving ability, whereas sequential learning benchmarks additionally examine whether experience improves performance on later tasks.

One line of evaluation studies experience reuse through sequential tasks and feedback. StreamBench retrieves feedback-selected examples for later tasks and tests robustness across five shuffled task orders \cite{wu2024streambench}. LifelongAgentBench links database, operating-system, and knowledge-graph tasks through shared atomic skills and evaluates replay of successful trajectories in a fixed sequence \cite{zheng2025lifelongagentbench}. Its task environments are reset between tasks, while prior trajectories can be retained as experience. In \referencehighlight{Table~\ref{tab:benchmark-comparison}}, shared skills therefore count as procedural reuse rather than required dependencies on earlier task outputs; a fixed sequence alone does not count as controlled variation of task order. SEA-Eval employs repeated and orthogonal tasks to gauge local gain and stability \cite{jiang2026seaeval}. Other examples include GDPevo and BenchTrace, which inject controlled success, failure, and noise signals for reflection analysis \cite{zhou2026gdpevo, jiang2026benchtrace}. These settings examine learning from prior experience, while EvoLink-Bench also includes downstream tasks that require earlier facts, decisions, or artifacts.
To capture broader temporal dynamics, a second approach emphasizes stream length and cross-domain interference. Evo-Memory and AgentStream create heterogeneous streams with a shared evolutionary state, enabling the study of accumulation, forgetting, and domain switching \cite{wei2025evomemory, yan2026agentstream}. FinEvo-Bench investigates procedural experience reuse across financial cases using interleaved streams and paired non-evolving controls \cite{deng2026finevobench}. Recent domain-specific benchmarks like EvoPathBench and ScienceClaw-Eval further examine the evolution process by evaluating persistent program improvements and rule adaptations across scientific disciplines \cite{lin2026evopathbench, zhang2026scienceclaw}. However, the constituent tasks in these benchmarks remain largely independent; while an earlier episode might provide a useful strategy, a later request is generally still solvable even if the earlier episode is entirely removed.
Moving beyond sequence length and domain shifts, a third body of work focuses on environmental realism by evaluating persistent states across interactive sessions. AppWorld, $\tau$-bench, BFCL Multi-Turn, and RealUserSim ground agents in executable tools, dynamic states, and user feedback \cite{trivedi2024appworld, yao2024taubench, berkeley2024bfcl, salesforce2026realusersim}. Frameworks like ClawArena, CL-bench Life, StuLife, and Trainee-Bench couple decisions through long timelines, schedules, and hidden information \cite{ji2026clawarena, dou2026clbenchlife, cai2025ell, fu2026traineebench}. EvoArena introduces versioned software and preference chains \cite{xu2026evoarena}, while PAST-Bench tests cross-session memory and procedural reuse through matched persistence controls \cite{xue2026pastbench}. These settings successfully demonstrate the value of persistent states, yet their temporal structures are largely dictated by scenario-specific sequences rather than explicit, cross-task evolutionary dependencies. 

EvoLink-Bench builds upon these foundations by including interdependent streams in which historical context is necessary for specific downstream tasks, alongside controlled variations in task order, spacing, and interference. As highlighted in \referencehighlight{Table~\ref{tab:benchmark-comparison}}, EvoLink-Bench uniquely combines cross-task dependencies, controlled scheduling, knowledge/rule revision, and procedural experience reuse. This combination enables a rigorous evaluation of whether agents can effectively revise and reuse experience as interdependent tasks arrive under conditions of delay and interference.

\section{Problem Formulation: Persistent Streaming Self-Evolution}
\label{sec:problem}

Existing self-evolution benchmarks often evaluate tasks independently or process each stream item only once, abstracting away the persistent and iterative dynamics of real-world service~\cite{wu2024streambench,jiang2026seaeval,yan2026agentstream}. To more faithfully capture these dynamics in realistic deployment, our formulation models self-evolution as continual service over a long-horizon task stream $\mathcal{Q}=(q_1,\ldots,q_T)$. Each stream may pose a different type of problem, such as a user request, an autonomous-workflow objective, or an environment-triggered event. We curate task order and arrival patterns so that later items may depend on earlier outputs and both task occurrence and dependency structures vary across streams, exposing agents to realistic temporal dynamics and inter-task dependencies. Processing $q_t$ defines a stream turn, within a turn, the agent may make several attempts, each comprising multiple interaction steps, before the stream advances. Retrying $q_t$ is within-turn adaptation, whereas improved behavior on later items constitutes cross-turn self-evolution. Within-turn attempts share state rather than replay the initial condition.

\subsection{Continuous Service Process}
We formulate a continuous service process as the four-tuple
\begin{equation}
    \mathcal{I}=(\pi_{\theta},\mathcal{H},C,W),
    \label{eq:continuous-service-process}
\end{equation}
where $\pi_{\theta}$ is a parametric LLM agent that outputs a language sequence $a$ conditioned on an input context $c$, i.e., $a\sim\pi_{\theta}(\cdot\mid c)$.
$\mathcal{H}$ is the harness that constructs model inputs, interprets language actions, and mediates tool execution. $C$ denotes the persistent model context, including the active interaction history and textual experience available to the harness. $W$ denotes the persistent external state, including the workspace, system state, and relevant world state.

At turn $t$, the current instance of the service process is
\begin{equation}
    \mathcal{I}_t=(\pi_{\theta_t},\mathcal{H}_t,C_t,W_t).
    \label{eq:turn-configuration}
\end{equation}
Serving $q_t$ under $\mathcal{I}_t$ produces a turn-level interaction trajectory $\tau_t$ and a turn-level reward $r_t$,
\begin{equation}
    (\tau_t,r_t)=\operatorname{Execute}_t(q_t;\mathcal{I}_t).
    \label{eq:turn-execution}
\end{equation}
The actions recorded in $\tau_t$ operate on the live external state, transforming $W_t$ into $W_{t+1}$. After execution, an associated update algorithm $U_t$ performs self-evolution by revising the model, harness, or persistent context via $\tau_t$ and $r_t$,
\begin{equation}
    (\pi_{\theta_{t+1}},\mathcal{H}_{t+1},C_{t+1})
    =U_t(\pi_{\theta_t},\mathcal{H}_t,C_t,\tau_t,r_t).
    \label{eq:turn-update}
\end{equation}
Here, $U_t$ is an agent-defined update function whose concrete form depends on the agent implementation. It may accumulate memory in $C_{t+1}$ by reflecting on the completed trajectory $\tau_t$ and its reward $r_t$, update the model parameters $\theta_t$ through reinforcement learning, or combine both mechanisms.
The next turn therefore begins from $\mathcal{I}_{t+1}=(\pi_{\theta_{t+1}},\mathcal{H}_{t+1},C_{t+1},W_{t+1})$. When no behavioral update is applied, $U_t$ leaves $\pi_{\theta_t}$ and $\mathcal{H}_t$ unchanged and derives $C_{t+1}$ only from $C_t$ and $\tau_t$. Across the complete service process, our benchmark records the reward sequence $\mathcal{R}=(r_1,\ldots,r_T)$ and uses its trajectory over the stream to measure the agent's self-evolution performance.

\subsection{Within-Turn Execution}
\label{sec:within-turn}

For each question $q_t$, execution produces an alternating sequence of agent actions and environment responses,
\begin{equation}
    \tau_t=(a^t_1,s^t_1,a^t_2,s^t_2,\ldots,a^t_{n_t},s^t_{n_t}),
    \label{eq:turn-trajectory}
\end{equation}
where $a^t_i$ is an action and $s^t_i$ is the corresponding response from the environment. An action may invoke a tool, request clarification, or submit a final response to the question. For a tool call, the harness executes the requested tool and returns its result as $s^t_i$. For a clarification request, the user or environment returns the requested information. For a submitted response, the environment evaluates the answer and returns feedback together with a score.

To reflect the iterative nature of real-world service, the evaluation framework allows the agent to revise a low-scoring response within the same turn. The resulting feedback is appended to the trajectory, after which the agent may perform additional tool calls, request further clarification, and submit another response. Consequently, $\tau_t$ may contain multiple submitted responses and their associated feedback. The turn ends when a response is accepted or the maximum number of response submissions is reached, and the score of the last submitted response defines the turn-level reward $r_t$.

\begin{figure}[H]
    \centering
    \IfFileExists{figures/Within_Turn_Trajectory.pdf}{%
        \includegraphics[width=\linewidth]{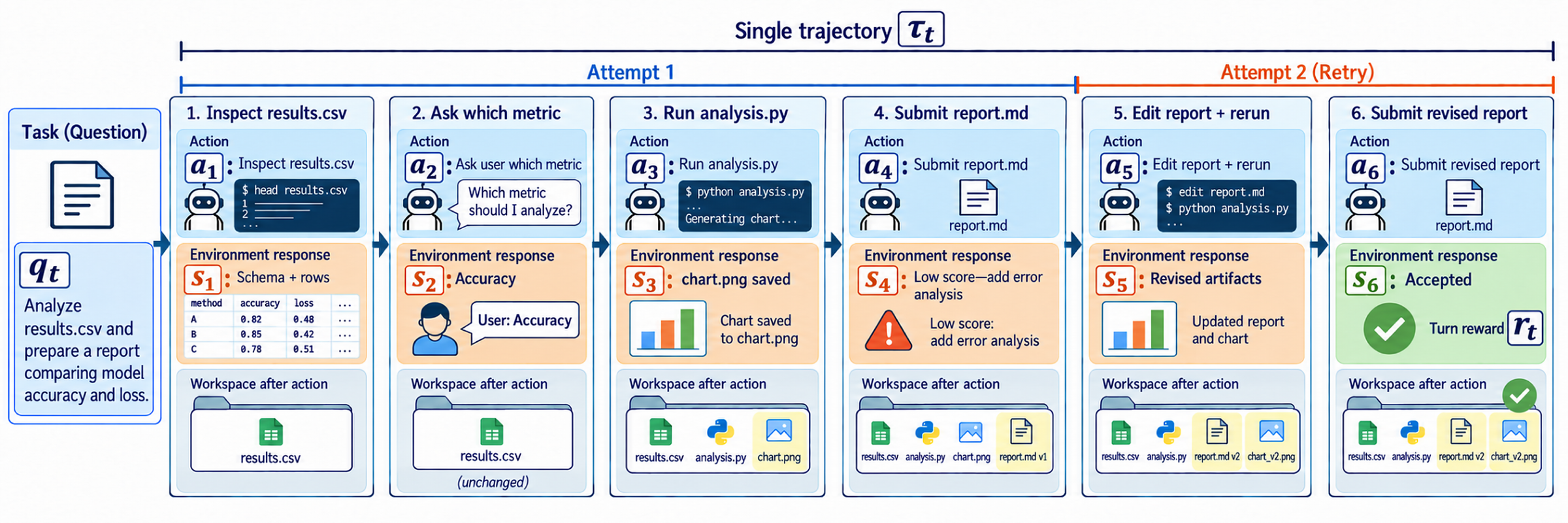}%
    }{%
        \fbox{\parbox{0.94\linewidth}{\centering
        \referencehighlight{Figure~\ref{fig:within-turn-execution}} is available in the full project build.}}%
    }
    \caption{An illustrative within-turn trajectory. The agent inspects data, requests clarification, runs analysis tools, and writes a report, while each action adds to or modifies the same persistent workspace. Feedback on the initial report triggers a stateful retry that produces revised artifacts before the turn terminates with reward $r_t$.}
    \label{fig:within-turn-execution}
\end{figure}

A visual example of this within-turn execution process is shown in \referencehighlight{Figure~\ref{fig:within-turn-execution}}.

\section{Hierarchical Evolution Task Construction}
\label{sec:task-construction}

To diagnose what an agent actually improves during continual service, we separate self-evolution into two complementary levels. \textbf{L1 (language-expressible factual evolution)} evaluates whether an agent can acquire, retain, and reuse explicit knowledge, including facts, preferences, constraints, and tool conventions. \textbf{L2 (task-level capability evolution)} evaluates whether the agent can improve reusable problem-solving behavior through execution and feedback. This separation reveals whether a method primarily strengthens memory and retrieval, task-solving competence, or both.

Each level covers multiple task categories and stream structures. Tasks of the same category may occur in consecutive turns or recur after several intervening turns, and later tasks may depend on information, artifacts, or experience produced earlier in the stream. Together, these variations test immediate acquisition, long-term retention, transfer, and dependency-aware reuse under realistic arrival patterns. 
The temporal controls also address a challenge identified in long-context evaluation: information being present in the input does not ensure that it is used reliably. Controlled retrieval and question-answering experiments found sensitivity to the position of relevant information \cite{liu2024lostmiddle}. We therefore vary spacing and intervening work explicitly; these are diagnostic conditions, rather than an assumption that the same positional effect must occur in every evaluated agent.

\subsection{L1 Language-Expressible Factual Evolution}
\label{sec:l1-construction}

\paragraph{Task construction.}
L1 scenario tasks are drawn from the real-source-grounded release of ClawArena~\cite{ji2026clawarena}. We retain the original task instructions, workspace evidence, conversation histories, staged updates, and executable success criteria. Each scenario defines a user role and a persistent information environment containing documents and structured records. Requests require evidence retrieval, conflict resolution, classification, or artifact production under scenario facts, user preferences, and standing constraints. Authoritative records coexist with incomplete summaries and obsolete statements, requiring the agent to identify relevant evidence and determine which statements remain valid. The original executable checks assess the resulting artifacts, with separate checks for preference adherence where specified.

We add domain-specific knowledge threads centered on procedural rules, including evidence-grading standards and reporting conventions. Each rule specifies conditions of use and is disclosed before its applications. Application requests provide cases and required outputs without restating the rule. Case sets vary evidence types, numerical thresholds, and exception conditions. For example, an evidence-grading request presents findings supported by a reproducible demonstration, a vendor advisory, an unverified claim, or static analysis; the agent must assign grades using the previously disclosed standard. Further applications vary requests or extend earlier outputs under the same rule.

Each task specifies accessible evidence, permitted actions, and criteria for success. We verify evidence availability, the consistency of environmental updates, and the order of disclosure and application. Knowledge-thread tasks are evaluated for rule application, incorporation of corrections, recall accuracy, and output completeness. Reference specifications and evaluation criteria are withheld from the agent; feedback on unsuccessful attempts is limited to observable deficiencies.

The distinction between corrected rules, temporary exceptions, and unavailable information mirrors LongMemEval’s separate evaluation of knowledge updates and abstention \cite{wu2025longmemeval}. Here, these checks are embedded in executable scenario requests, so evaluation focuses on whether the agent applies the appropriate rule in its output.

\begin{figure}[H]
\centering
\includegraphics[width=\linewidth]{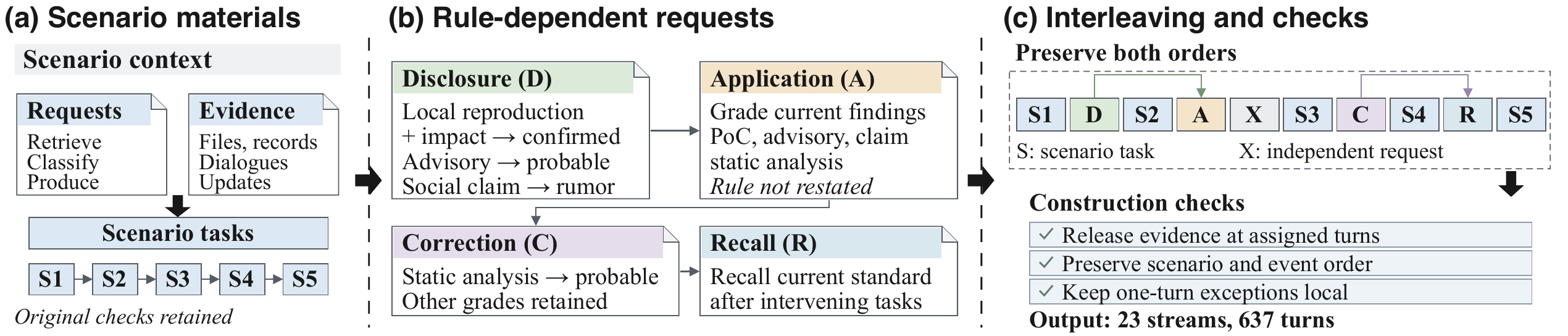}
\captionsetup{font=small,labelfont=bf,skip=4pt}
\caption{L1 construction. Preserved scenario materials are augmented with rule-dependent requests. An evidence-grading thread illustrates disclosure, application without rule restatement, correction, and optional recall. Interleaving preserves both orders.}
\label{fig:l1-construction}
\end{figure}

\paragraph{Stream construction.}
We construct L1 streams by inserting knowledge-thread events and independent distractors into an ordered sequence of scenario requests (\referencehighlight{Figure~\ref{fig:l1-construction}}). Conversation history and the information environment persist across turns, so later requests can depend on prior decisions and artifacts. New evidence becomes accessible only at designated turns. We represent knowledge thread $j$ by the partial order $\mathcal{P}_j=(\mathcal{V}_j,\prec_j)$ over disclosure, applications, correction, optional recall, and diagnostic events. An injective schedule assigns events to turns subject to their precedence constraints:
\begin{equation}
    \sigma:\bigcup_j\mathcal{V}_j\hookrightarrow\{1,\ldots,T\},
    \qquad
    u\prec_j v\;\Longrightarrow\;\sigma(u)<\sigma(v).
    \label{eq:l1-event-schedule}
\end{equation}
Scenario tasks and distractors fill the other positions without altering scenario order.

Disclosure establishes a standing rule for later application requests, which supply case data without restating the rule. Correction follows the first application and precedes any recall; additional applications may occur before or after correction. A correction revises the affected clauses and preserves those left unchanged. Applications and recall after correction must use the amended rule. Temporary exceptions apply solely to the current request and leave the standing rule unchanged. A probe preceding disclosure requires acknowledgement that the governing rule is unavailable.

The six schedules control the spacing and recurrence of knowledge events. Adjacent schedules place disclosure and first application in consecutive turns, while dispersed schedules separate knowledge events with scenario work. Long-gap schedules combine early disclosure with late recall; repeated-application schedules reuse the rule on several requests. The remaining schedules place a probe before disclosure or a temporary exception between correction and recall. Retention intervals count intervening turns from disclosure or correction to recall, distinguishing the age of the initial rule from that of its amendment. Independent questions and formatting exceptions add interference without permitting persistent changes to standing rules or unrelated artifacts.

\begin{figure}[!t]
\centering
\captionsetup{font=small,labelfont=bf,skip=5pt}
\begin{minipage}[t]{0.485\linewidth}
\vspace{0pt}
\includegraphics[width=\linewidth]{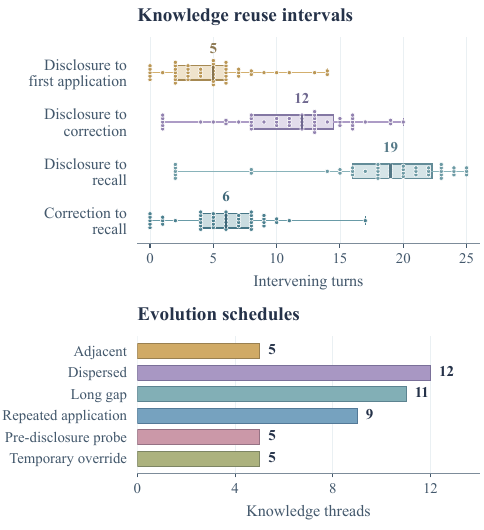}
\caption{L1 temporal structure. Upper: event intervals across 47 threads (44 with recall). Dots denote threads; boxes show interquartile ranges and medians, and whiskers span observed ranges. Lower: coverage of six evolution schedules.}
\label{fig:l1-temporal-structure}
\end{minipage}\hfill
\begin{minipage}[t]{0.485\linewidth}
\vspace{0pt}
\includegraphics[width=\linewidth]{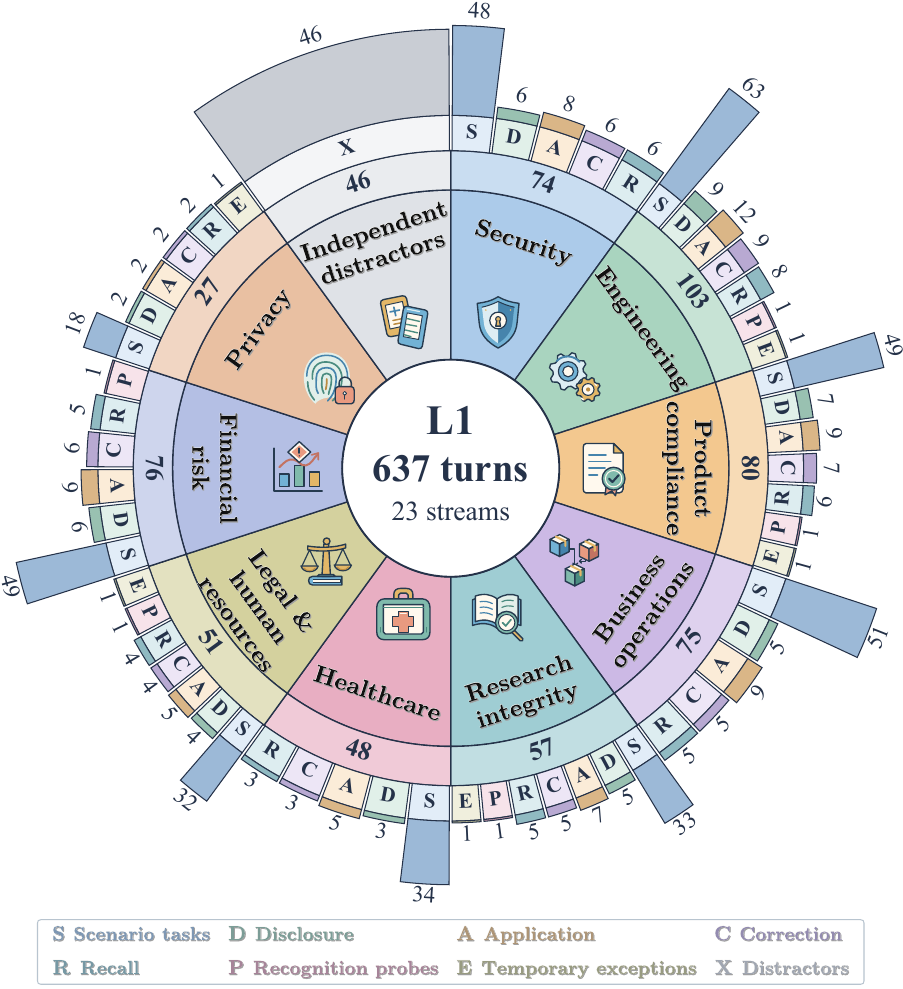}
\caption{L1 task composition: 637 turns across nine domains and a distractor category. Inner sectors identify domains and their totals; colored outer bars distinguish task groups. Sector angles are fixed. Bar lengths and labels encode turn counts.}
\label{fig:l1-composition}
\end{minipage}
\end{figure}

\paragraph{Statistics.}
L1 comprises 637 turns across 23 persistent streams in nine domains, with 23--32 turns per stream. \referencehighlight{Figures~\ref{fig:l1-temporal-structure} and~\ref{fig:l1-composition}} summarize its temporal structure and task composition. Scenario tasks account for most turns, while knowledge threads contribute approximately one-third. Rule disclosure and application occur within ongoing scenario work, where later decisions can depend on earlier instructions, evidence, and artifacts. Standing preferences constrain successive requests, and new evidence can require earlier conclusions to be revised. Independent distractors add unrelated questions and temporary exceptions whose effects must remain confined to the current request.

The 47 knowledge threads span all six evolution schedules and contain 66 application turns, reflecting repeated use of some rules. Dispersed and long-gap schedules are the most common, making reuse after intervening work a recurring requirement. Pre-disclosure probes test recognition of missing rules; temporary overrides test whether exceptions remain local. \referencehighlight{Figure~\ref{fig:l1-temporal-structure}} shows that first application generally follows disclosure after several intervening requests, while recall occurs later. Among the 44 threads with recall, the median disclosure-to-recall interval is 19 intervening turns, reaching 25 in the longest case; the median correction-to-recall interval is six turns. Recall must preserve unchanged clauses from the initial rule and incorporate subsequent amendments.

\subsection{L2 Task-Level Capability Evolution}
\label{sec:l2-construction}
L2 asks whether experience from earlier tasks improves the agent on a new task that requires the same capability. Each capability is practiced on several tasks and then measured on a \emph{probe}, a final unseen task of that capability. A stream controls what lies between practice and probe along four axes: how many turns intervene, which work fills them, whether the probe moves to another benchmark, and whether a later task must build on files that an earlier task produced. Machine learning engineering (MLE) tasks are central to L2 because they require an experimental loop of data diagnosis, model choice, and tuning under a compute budget. MLE-bench similarly evaluates dataset preparation, model training, and experimentation as components of ML engineering \cite{chan2024mlebench}; our MLE tasks are drawn from AIRS-Bench and MLGym, as specified below.

\paragraph{Tasks and capabilities.}
Every turn is a native task from SRA-Bench~\cite{su2026srabench}, $\tau^2$-Bench~\cite{barres2025tau2bench}, SkillLearnBench~\cite{zhong2026skilllearnbench}, SkillsBench~\cite{li2026skillsbench}, AIRS-Bench~\cite{lupidi2026airsbench}, or MLGym~\cite{nathani2025mlgym}, with its original instructions, inputs, and evaluation. Programming tasks are scored by unit tests and document and data-analysis tasks by the upstream tests on the produced files. Service tasks run in the original $\tau^2$ environment: the agent calls its tools and converses with its user simulator, and the final database state and dialogue are checked against the task's assertions. A task is admitted only if its upstream reference solution passes this evaluation and a corrupted submission fails in an offline container. Each task carries a capability label for the solution procedure it shares with other tasks, derived from source annotations such as SkillLearnBench task families, SkillsBench categories, $\tau^2$ workflow types, the libraries exercised by SRA tasks, and the problem type of MLE tasks. Tasks of one capability differ in inputs and required outputs, so a probe rewards the procedure rather than a remembered answer. Reference solutions, tests, and hidden labels remain outside the agent's environment.

MLE turns use distinct AIRS-Bench and MLGym tasks, each evaluated on its full hidden test set within the upstream time budget. Their turn score is the AIRS-Bench normalized score, which maps the native metric to 0 at the task's worst or baseline anchor and to 1 at the best reported result, so partial progress below the state of the art remains visible. Because CPU and GPU tasks demand different experimental practice, they form separate streams: one with CPU-feasible tasks and one dominated by GPU-class tasks.

\begin{figure}[!tbp]
	\centering
	\includegraphics[width=\linewidth]{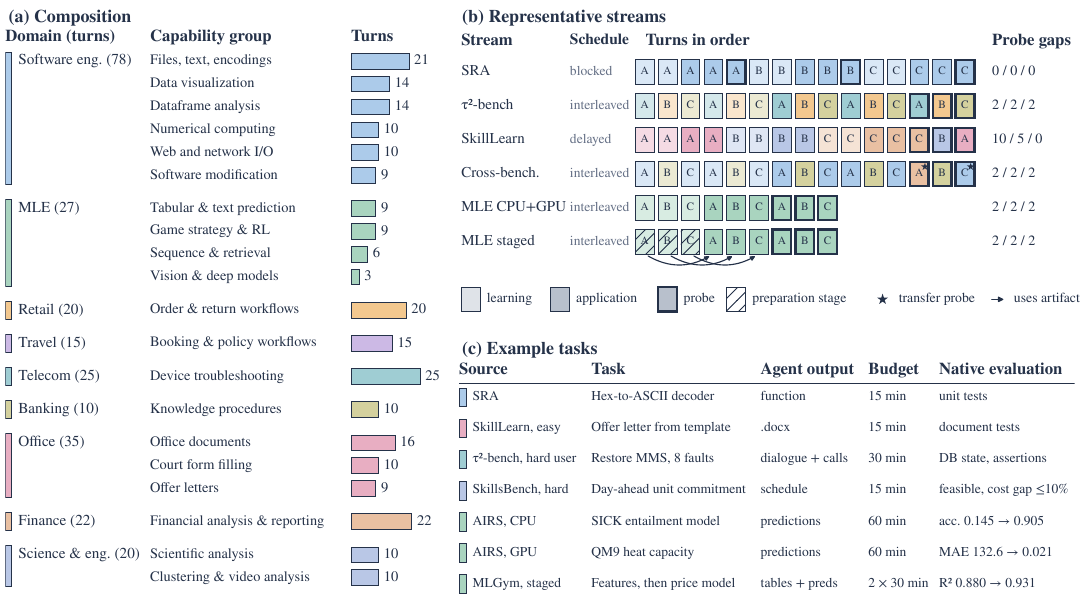}
	\caption{L2 composition. (a) Turns per capability group, grouped by domain. (b) One stream of each kind; cells are turns in order, colored by domain and lettered by capability, with probe gaps $g_c$ for capabilities A/B/C. (c) Example tasks with their outputs, time budgets, and native evaluation; MLE arrows mark metric values mapped to 0 and 1.}
	\label{fig:l2-composition}
\end{figure}

\paragraph{Stream construction.}
Each stream combines three capabilities with $n$ distinct tasks each ($n=5$; $n=3$ in MLE streams). The first two tasks of a capability are \emph{learning} turns (the first task when $n=3$), the last is the probe, and the rest are \emph{application} turns; the roles share the same evaluation and differ only in position. All turns of a stream share one persistent workspace in which each task adds its own directory and earlier files remain available.

For capability $c$ whose tasks occupy turns $\sigma(c,1)<\cdots<\sigma(c,n)$, the probe gap counts the turns between the probe and the previous task of the same capability:
\begin{equation}
    g_c=\sigma(c,n)-\sigma(c,n-1)-1 .
    \label{eq:l2-probe-gap}
\end{equation}
\emph{Blocked} streams place each capability contiguously ($g_c=0$) and test immediate reuse. \emph{Interleaved} streams rotate the capabilities ($g_c=2$) and test reuse under interference. \emph{Delayed} streams keep the stream length but defer the probes to the end in reverse order ($g_c\in\{0,5,10\}$) and test retention over increasing intervals. The intervening work is either the same benchmark's other capabilities or, in cross-benchmark streams, tasks with different tools, environments, and output formats. In two cross-benchmark streams, a probe comes from another benchmark whose capability shares the procedure, for example a SkillsBench financial-data task after SRA dataframe-analysis tasks, which tests transfer.

The dependency stream makes later MLE work depend on earlier artifacts. It splits three MLE tasks into a preparation stage and the original task: building feature tables before house-price regression, cleaning a time series before forecasting, and implementing a payoff-estimation harness before designing a Colonel Blotto strategy. The original task starts in an empty directory; its data and the prepared artifact exist only in the preparation stage's directory, so it cannot proceed without that earlier turn. Reference checks assess preparation; unseen probes end each line.

\paragraph{Statistics.}
\newcommand{\LTwoTurnCount}{252}
\newcommand{\LTwoStreamCount}{18}
\newcommand{\LTwoUniqueSourceCount}{181}
\newcommand{\LTwoDomainCount}{9}
\newcommand{\LTwoCapabilityCount}{20}
\newcommand{\LTwoCodebookCapabilityCount}{33}
\newcommand{\LTwoSourceCount}{6}
\newcommand{\LTwoAutoResearchTurnCount}{27}
\newcommand{\LTwoProbeTurnCount}{54}

L2 comprises \LTwoTurnCount{} turns in \LTwoStreamCount{} streams, drawn from \LTwoUniqueSourceCount{} distinct tasks of \LTwoSourceCount{} benchmarks and spanning \LTwoDomainCount{} domains and \LTwoCapabilityCount{} capability groups (\referencehighlight{Figure~\ref{fig:l2-composition}a}). Ten intra-benchmark and five cross-benchmark streams have 15 turns each; the CPU, CPU+GPU, and dependency streams have nine turns each and contribute \LTwoAutoResearchTurnCount{} MLE turns. Of the \LTwoProbeTurnCount{} probes, 18 directly follow the previous task of their capability, 24 follow two intervening tasks, and 12 follow five or ten; four probes test cross-benchmark transfer. \referencehighlight{Figure~\ref{fig:l2-composition}b} shows one stream of each kind, and \referencehighlight{Figure~\ref{fig:l2-composition}c} shows that tasks range from a single function checked by unit tests to hour-long model training scored against the best reported result.

\section{Experiments}
\label{sec:experiments}

We conduct a systematic evaluation of five agent frameworks across three backbone configurations and three persistence conditions, guided by the following five research questions:
\begin{enumerate}[leftmargin=*]
    \item[\ding{182}] \textbf{\textit{RQ1:}} How do the evaluated frameworks perform across the two evolution levels, L1 and L2?
    \item[\ding{183}] \textbf{\textit{RQ2:}} How are retained state and post-turn updates related to task performance?
    \item[\ding{184}] \textbf{\textit{RQ3:}} How much can agents recover from an unsuccessful submission through within-task retries?
    \item[\ding{185}] \textbf{\textit{RQ4:}} Do recorded changes to persistent state correspond to better performance on later probe tasks?
    \item[\ding{186}] \textbf{\textit{RQ5:}} How do the observed results vary across task domains and independent streams?
\end{enumerate}

\subsection{Experimental Setup}
\label{sec:experimental-setup}

\paragraph{Backbones and agent frameworks.}
We evaluate three backbone models: DeepSeek V4 Flash \cite{deepseek2026v4modelcard}, GLM-5.3 \cite{zai2026glm53modelcard}, and GPT-6.1 Terra \cite{openai2026gpt61terra}. DeepSeek V4 Flash and GLM-5.3 are deployed locally, while GPT-6.1 Terra is accessed through its API.
Four agent frameworks are evaluated across all three configurations: Hermes \cite{nous2026hermesagent}, Prime Agent \cite{karten2026primeagent}, Adaptive Auto-Harness \cite{liu2026adaptiveharness}, and OpenCode \cite{anomaly2026opencode}. The fifth framework is configuration-specific: DeepSeek Harness \cite{deepseek2026harness} for DeepSeek V4 Flash, ZCode \cite{zai2026zcode} for GLM-5.3, and Codex \cite{openai2026codexrepo} for GPT-6.1 Terra. These frameworks are evaluated within their respective configurations and are not directly interchangeable. Among the evaluated frameworks, Hermes reviews and updates memory and skills, Prime Agent refines a persistent harness, and Adaptive Auto-Harness updates prompts, skills, and memory. In the audited DeepSeek V4 Flash and GLM-5.3 configurations, OpenCode, DeepSeek Harness, and ZCode do not have an enabled native post-turn learner.

\begin{figure}[!htbp]
	\centering
	\includegraphics[width=\linewidth]{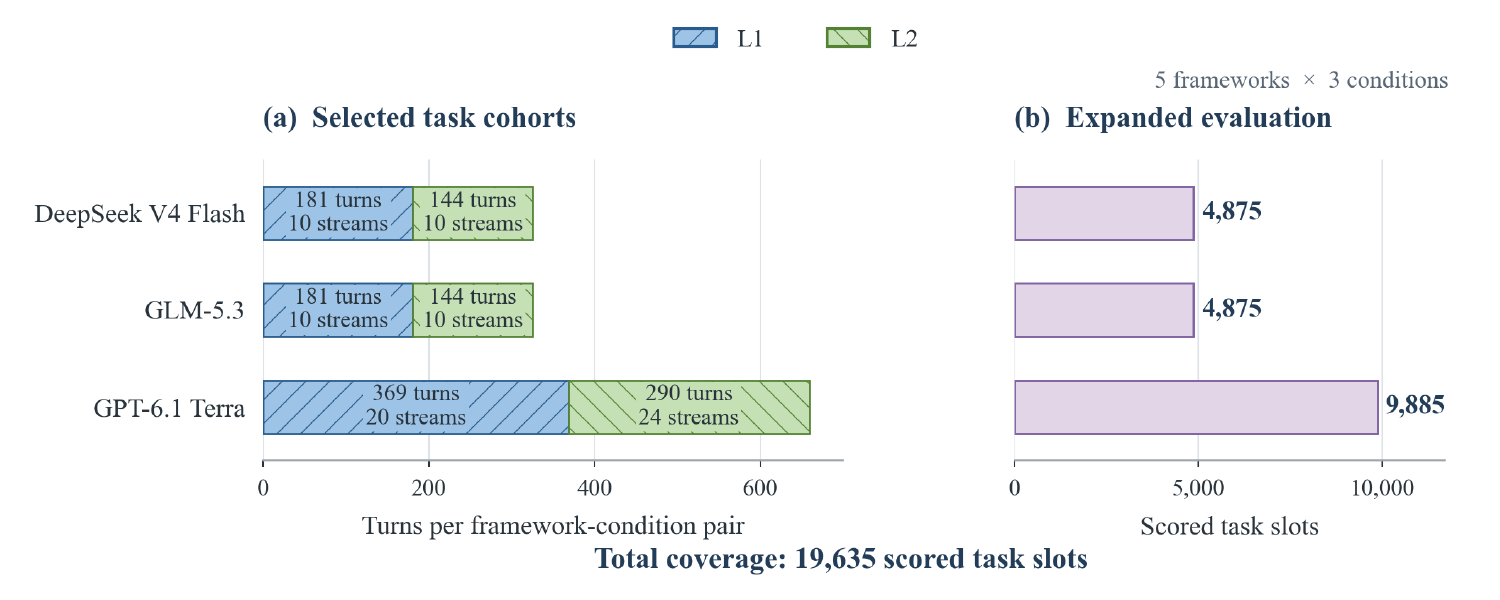}
	\caption{Evaluation coverage. Left: L1 and L2 turn counts and stream coverage for each framework--condition pair. Right: total scored task slots across five frameworks and three persistence conditions. The 19,635 slots represent evaluated task positions rather than unique source tasks.}
	\label{fig:evaluation-coverage}
\end{figure}

\paragraph{Evaluation cohorts and coverage.}
Due to limited model-call budgets and execution time, we evaluate selected task cohorts rather than running every agent framework on all benchmark streams. The evaluated cohorts also differ across backbone configurations. For DeepSeek V4 Flash and GLM-5.3, the selected L1 panel contains 181 turns across 10 streams, while the L2 panel contains 144 turns across 10 streams. The L2 panel consists of 135 turns from the primary intra-benchmark panel and nine CPU AutoResearch turns. For GPT-6.1 Terra, each framework--condition pair covers 369 L1 turns across 20 streams and 290 L2 turns across 24 streams. Figure~\ref{fig:evaluation-coverage} summarizes the evaluation coverage. Across the five frameworks and three persistence conditions, the selected cohorts account for 19,635 scored task slots. Each slot represents one task position evaluated under a particular framework and condition.

The selected DeepSeek V4 Flash and GLM-5.3 cohorts do not include the dedicated artifact-dependency, GPU, and cross-benchmark L2 streams described in the benchmark construction. In contrast, the GPT-6.1 Terra evaluation covers a larger set of L1 and L2 tasks. Therefore, the current experiments do not provide uniform coverage of all benchmark stream types, and the results should be interpreted within the scope of the evaluated cohorts. We report each configuration separately; differences are not controlled backbone comparisons.

\paragraph{Persistence conditions.}
\textbf{Reset} reinitializes the workspace and agent state before each turn and disables explicit post-turn updates. \textbf{Frozen} retains the permitted workspace and agent state but disables those updates. \textbf{Evolving} retains state and enables the post-turn update procedure when the evaluated framework implements one. Thus, Evolving describes an execution condition; it does not guarantee that an update was attempted, committed, or useful.

For an aggregate score $M$, we report two descriptive contrasts:
\begin{equation}
    \Delta_{\mathrm{persist}}=M_{\mathrm{Frozen}}-M_{\mathrm{Reset}},
    \qquad
    \Delta_{E-F}=M_{\mathrm{Evolving}}-M_{\mathrm{Frozen}}.
    \label{eq:evolution-gains}
\end{equation}
The first contrast captures the observed difference associated with retaining state. The second compares the Evolving and Frozen conditions. Neither contrast is interpreted as a causal estimate of a learning algorithm, particularly when the evaluated integration has no native learner or when runtime revisions and budgets differ.

\paragraph{Scoring and reporting.}
A turn's reward $r_{s,t}\in[0,1]$ differs from its pass/fail result. We use the last recorded submission, keep partial rewards, and set rewards to zero only for recorded execution errors or timeouts. The main score is the mean reward across evaluated turns, multiplied by 100. We also report submission pass rates, paired stream differences, saved-state changes, and final probe scores. Appendix~\ref{app:experimental-details} gives scoring and state rules.

\paragraph{Runtime provenance.}
The DeepSeek V4 Flash and GLM-5.3 selections include retained and replacement runs across several runtime revisions. The replacement scope was fixed before inspecting replacement scores, and all displayed results were recomputed from the selected terminal records. However, the historical and replacement runs do not form a certified equal-budget cohort, and repeated-run uncertainty is not available. We therefore treat the results as descriptive benchmark observations rather than statistically controlled estimates.

This reporting boundary follows the methodological concern that agent accuracy should be assessed together with execution cost and standardized evaluation conditions \cite{kapoor2024agentsmatter}. MLE-bench also studies resource scaling explicitly \cite{chan2024mlebench}. Accordingly, the condition contrasts below describe the selected runs and do not establish which framework would lead under a common compute or model-call budget.

\subsection{Main Experimental Results (RQ1)}
\label{sec:rq1-overall}

We first compare the five agent frameworks on L1 and L2 to examine whether their strengths hold across task levels and persistence conditions. \referencehighlight{Tables~\ref{tab:agent-results}--\ref{tab:agent-results-terra}} report the mean final reward within each backbone configuration.

\begin{table}[!htbp]
    \centering
    \caption{DeepSeek V4 Flash mean rewards (0--100, $\uparrow$): 181 L1 / 144 L2 turns per framework--condition pair.}
    \label{tab:agent-results}
    \small
    \setlength{\tabcolsep}{8pt}
    \renewcommand{\arraystretch}{1.25}
    \begin{tabular}{lrrrrrr}
        \toprule
        \rowcolor{tableheader}
         & \multicolumn{3}{c}{\textbf{L1 (10 streams)}} & \multicolumn{3}{c}{\textbf{L2 (10 streams)}} \\
        \rowcolor{tableheader}
        \multirow{-2}{*}{\textbf{Framework}} & \textbf{Reset} & \textbf{Frozen} & \textbf{Evolving} & \textbf{Reset} & \textbf{Frozen} & \textbf{Evolving} \\
        \midrule[\heavyrulewidth]
        Hermes & 66.08 & 68.40 & 74.75 & 47.80 & 51.13 & 64.98 \\
        Prime Agent & 61.71 & 70.83 & 64.09 & 56.33 & 52.58 & 67.35 \\
        Adaptive Auto-Harness & 71.49 & 76.57 & \textbf{92.15} & 54.22 & 55.41 & 56.64 \\
        OpenCode & \textbf{81.38} & \textbf{86.80} & 89.12 & 59.34 & 58.30 & 64.30 \\
        DeepSeek Harness & 76.91 & 84.48 & 89.45 & \textbf{64.31} & \textbf{72.92} & \textbf{80.63} \\
        \bottomrule
    \end{tabular}
\end{table}

\begin{table}[!htbp]
    \centering
    \caption{GLM-5.3 results. The task sets and aggregation match \referencehighlight{Table~\ref{tab:agent-results}}; the fifth framework is ZCode. Explicit execution exceptions and timeouts contribute zero only to the affected turn.}
    \label{tab:agent-results-glm}
    \small
    \setlength{\tabcolsep}{8pt}
    \renewcommand{\arraystretch}{1.25}
    \begin{tabular}{lrrrrrr}
        \toprule
        \rowcolor{tableheader}
         & \multicolumn{3}{c}{\textbf{L1 (10 streams)}} & \multicolumn{3}{c}{\textbf{L2 (10 streams)}} \\
        \rowcolor{tableheader}
        \multirow{-2}{*}{\textbf{Framework}} & \textbf{Reset} & \textbf{Frozen} & \textbf{Evolving} & \textbf{Reset} & \textbf{Frozen} & \textbf{Evolving} \\
        \midrule[\heavyrulewidth]
        Hermes & \textbf{85.47} & 90.17 & 87.96 & 72.05 & 71.16 & 71.12 \\
        Prime Agent & 67.73 & 76.08 & 68.18 & 66.04 & 64.17 & 65.05 \\
        Adaptive Auto-Harness & 83.87 & 89.34 & 90.66 & 69.32 & 70.06 & 75.00 \\
        OpenCode & 84.59 & 84.97 & 88.45 & \textbf{75.57} & \textbf{76.12} & 72.22 \\
        ZCode & 82.43 & \textbf{94.03} & \textbf{94.25} & 73.20 & 71.34 & \textbf{77.29} \\
        \bottomrule
    \end{tabular}
\end{table}

\begin{table}[!htbp]
    \centering
    \caption{GPT-6.1 Terra mean final scores (0--100): 369 L1 and 290 L2 turns per framework--condition pair.}
    \label{tab:agent-results-terra}
    \small
    \setlength{\tabcolsep}{8pt}
    \renewcommand{\arraystretch}{1.25}
    \begin{tabular}{lrrrrrr}
        \toprule
        \rowcolor{tableheader}
         & \multicolumn{3}{c}{\textbf{L1 (20 streams)}} & \multicolumn{3}{c}{\textbf{L2 (24 streams)}} \\
        \rowcolor{tableheader}
        \multirow{-2}{*}{\textbf{Framework}} & \textbf{Reset} & \textbf{Frozen} & \textbf{Evolving} & \textbf{Reset} & \textbf{Frozen} & \textbf{Evolving} \\
        \midrule[\heavyrulewidth]
        Hermes & 51.00 & 56.18 & 56.53 & 48.69 & \textbf{57.15} & 50.67 \\
        Prime Agent & 52.36 & 62.95 & 85.91 & 35.57 & 37.71 & 37.15 \\
        Adaptive Auto-Harness & 77.51 & 90.68 & \textbf{87.64} & 50.75 & 50.84 & 58.49 \\
        OpenCode & 45.31 & 81.68 & 86.99 & \textbf{57.00} & 57.07 & \textbf{63.24} \\
        Codex & \textbf{81.19} & \textbf{90.98} & \textbf{87.64} & 56.56 & 56.18 & 59.64 \\
        \bottomrule
    \end{tabular}
\end{table}

The rankings differ between the two task levels. Under DeepSeek V4 Flash Evolving, Adaptive Auto-Harness leads L1 with 92.15, while DeepSeek Harness leads L2 with 80.63. A similar split appears under GPT-6.1 Terra Evolving: Codex and Adaptive tie on L1, while OpenCode leads L2. These changes matter because L1 tests the use of facts, preferences, and rules, whereas L2 also requires working code, data, and other task artifacts. Success at retaining information does not by itself establish that a framework can turn that information into a correct task solution. Prime Agent illustrates this gap on GPT-6.1 Terra, scoring 85.91 on L1 Evolving and 37.15 on L2 Evolving. Since the levels use different tasks, these scores cannot directly compare their difficulty. Claims of broader capability need evidence from both levels.

The rankings also change with the execution condition. With DeepSeek V4 Flash, OpenCode leads L1 under Reset and Frozen, but Adaptive Auto-Harness takes the lead under Evolving. With GLM-5.3, OpenCode leads L2 under Reset and Frozen, while ZCode leads under Evolving. Thus, a framework selected for strong initial execution need not be the one selected for performance with retained state and updates enabled. Conversely, leading under Evolving does not establish that post-turn learning produced that lead: the score also reflects the framework's existing execution ability, access to retained artifacts, and within-task retries. The next analysis examines these condition differences.

These differences also affect how a combined score should be read. In the GPT-6.1 Terra evaluation, L1 contributes 55.99\% of the scored positions and L2 contributes 44.01\%. A combined score consequently gives more weight to L1, and changing this mixture could change the ranking when frameworks have different strengths across levels. We therefore retain separate L1 and L2 results as the main comparison. Comparisons also remain within each backbone configuration, since the evaluated task cohorts and the fifth framework differ across configurations. The resulting rankings describe performance on the selected tasks and do not establish a universal framework or backbone ranking.

\subsection{Persistence and Update Analysis (RQ2)}
\label{sec:rq2-conditions}

\begin{figure}[!htbp]
    \centering
    \includegraphics[width=\linewidth]{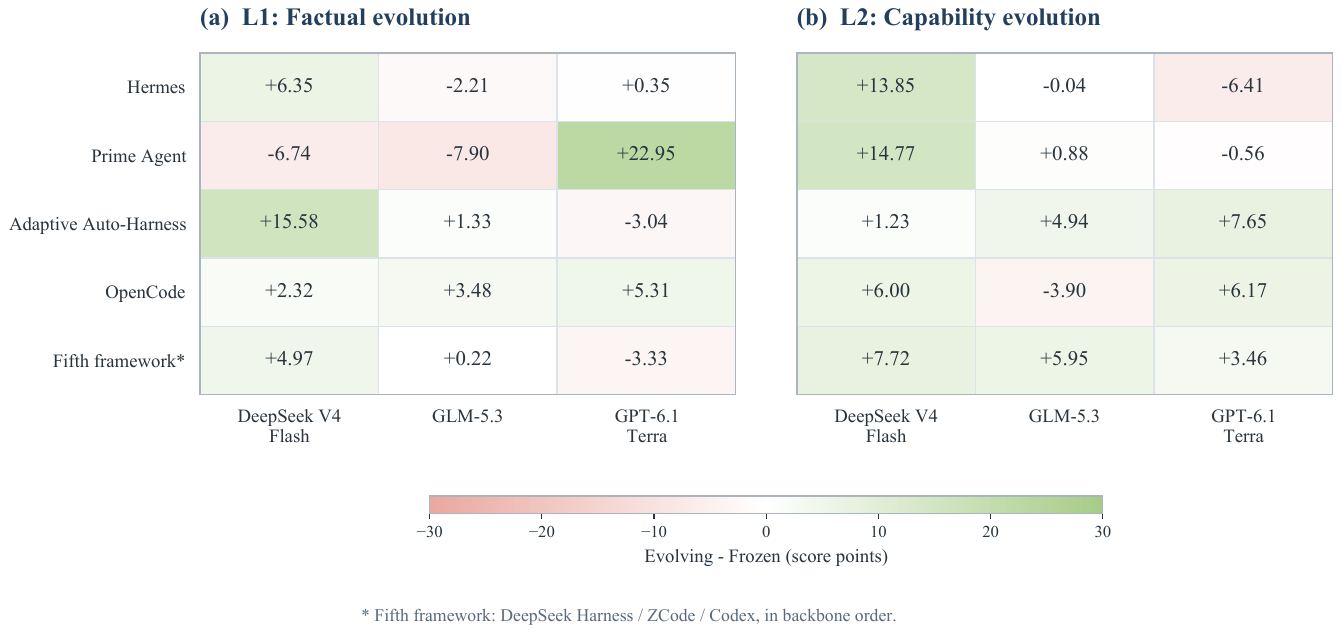}
    \caption{Observed Evolving--Frozen score differences across backbone configurations. Positive values indicate higher scores under Evolving; negative values indicate higher scores under Frozen. The fifth framework is configuration-specific: DeepSeek Harness, ZCode, and Codex, respectively. For DeepSeek V4 Flash and GLM-5.3, the fifth framework has no native post-turn learner; its contrast does not measure an update effect.}
    \label{fig:evolution-contrasts}
\end{figure}

We next examine how performance differs when earlier state is retained and when post-turn updates are enabled. Frozen--Reset compares runs with and without retained state, while Evolving--Frozen compares conditions with updates enabled when supported. \referencehighlight{Figure~\ref{fig:evolution-contrasts}} reports the latter difference. Reading the two comparisons together helps distinguish the value of retaining earlier work from the results observed when updates are also allowed.

\paragraph{Retaining state is not uniformly beneficial.}
On GPT-6.1 Terra L1, Frozen exceeds Reset for all five frameworks, with OpenCode showing the largest difference ($+36.37$ points). On L2, however, the differences for Adaptive, OpenCode, and Codex are close to zero ($+0.09$, $+0.07$, and $-0.38$). Retention therefore corresponds to substantial gains in one part of the benchmark without producing similar gains in another. The distinction is between making earlier information available and using it to solve later tasks: preserving files or context guarantees only the former. The scores do not reveal whether small gains arise from unused state, irrelevant content, or execution failures. They show that retention alone is insufficient evidence of effective reuse.

Enabling post-turn updates likewise does not produce a consistent advantage across task levels. Adaptive Auto-Harness has an Evolving--Frozen difference of $-3.04$ on GPT-6.1 Terra L1 but $+7.65$ on L2. Prime Agent shows the reverse pattern under DeepSeek V4 Flash, with $-6.74$ on L1 and $+14.77$ on L2. These sign changes within a configuration rule out a simple description in which the Evolving condition is uniformly better for an update-capable framework. An update procedure must be assessed against the tasks on which the saved content will be used, rather than judged only by whether it runs. The present comparisons identify where Evolving performs better or worse; they do not establish which update content explains a difference or whether it would persist across repeated runs.

The final average also hides changes within the stream. The cumulative results in Appendix~\ref{app:supplementary-quantitative} show that a condition's lead can change as more tasks are encountered. For DeepSeek V4 Flash Adaptive on L2, the Evolving--Frozen gap is $+4.30$ points at 25\% of the stream, $-4.68$ at 50\%, and $+1.23$ at the end. A positive final difference can therefore coexist with a disadvantage over an earlier part of the evaluation. Each progress point includes a different set of tasks, so these changes may reflect task composition as well as how the agent uses its history. Reporting intermediate results helps expose this variation, but the curves cannot by themselves establish steady learning or forgetting.

Interpreting these differences also requires checking whether an update actually occurs. The audited DeepSeek V4 Flash and GLM-5.3 integrations of OpenCode, DeepSeek Harness, and ZCode have no enabled native post-turn learner, so their Evolving--Frozen differences cannot demonstrate benefits from that learner. An enabled interface also does not guarantee a saved change. The condition comparisons therefore show where scores differ, while the state audit and probe analysis in Section~\ref{sec:rq4-reuse} examine the evidence for saved changes and later reuse.

\subsection{Within-Task Retry and Recovery Analysis (RQ3)}
\label{sec:rq3-retry}

To understand how agents use feedback within a task, we separate success on the initial submission from recovery over up to two retries. The pass/fail flag identifies when a task first passes, while the last recorded reward remains the task score. A submission may contain multiple model and tool calls. \referencehighlight{Table~\ref{tab:retry-outcomes-baselines}} pools L1, L2, and all three persistence conditions, giving 975 executions per framework. Each execution belongs to one outcome category, including a separate category for the 21 GLM-5.3 exception/timeout slots.

Self-Refine studies output revision using model-generated feedback \cite{madaan2023selfrefine}. Here, retries receive task-specific evaluator feedback. Their success measures the ability to correct the current solution, which is a different question from whether retained experience helps on a new task.

\begin{table}[!htbp]
    \centering
    \caption{Submission-level acceptance and non-acceptance rates (\%). Each framework has 975 task executions pooled over L1/L2 and Reset/Frozen/Evolving. ``No pass, 3'' means all three submissions were recorded without acceptance; ``No pass, $<3$'' means one or two submissions ended without acceptance. Exception/timeout slots are reported separately.}
    \label{tab:retry-outcomes-baselines}
    \small
    \setlength{\tabcolsep}{7pt}
    \renewcommand{\arraystretch}{1.25}
    \begin{tabular}{lrrrrrr}
        \toprule
        \rowcolor{tableheader}
        \multicolumn{1}{c}{\raisebox{0.5\normalbaselineskip}[0pt][0pt]{\textbf{Framework}}} & \textbf{Pass 1} & \textbf{Pass 2} & \textbf{Pass 3} & \shortstack{\textbf{No pass,}\\\textbf{3}} & \shortstack{\textbf{No pass,}\\$<3$} & \textbf{Exception} \\
        \midrule[\heavyrulewidth]
        \multicolumn{7}{l}{\textit{DeepSeek V4 Flash}} \\
        Hermes & 54.15\% & 4.72\% & 1.54\% & 16.41\% & 23.18\% & 0.00\% \\
        Prime Agent & 57.74\% & 3.08\% & 0.62\% & 9.64\% & 28.92\% & 0.00\% \\
        Adaptive Auto-Harness & 59.59\% & 5.23\% & 1.95\% & 23.08\% & 10.15\% & 0.00\% \\
        OpenCode & 64.10\% & 3.49\% & 1.44\% & 24.72\% & 6.26\% & 0.00\% \\
        DeepSeek Harness & 62.87\% & 7.90\% & 3.59\% & 19.69\% & 5.95\% & 0.00\% \\
        \midrule
        \multicolumn{7}{l}{\textit{GLM-5.3}} \\
        Hermes & 68.82\% & 6.87\% & 1.13\% & 14.36\% & 8.72\% & 0.10\% \\
        Prime Agent & 61.95\% & 2.97\% & 1.13\% & 9.13\% & 24.62\% & 0.21\% \\
        Adaptive Auto-Harness & 68.10\% & 4.92\% & 2.67\% & 16.62\% & 6.87\% & 0.82\% \\
        OpenCode & 67.38\% & 4.41\% & 1.74\% & 22.46\% & 4.00\% & 0.00\% \\
        ZCode & 76.41\% & 2.87\% & 1.44\% & 10.87\% & 7.38\% & 1.03\% \\
        \bottomrule
    \end{tabular}
\end{table}

\paragraph{Most success comes from the first submission.}
Across all 4,875 executions per backbone, first-submission acceptance is 59.69\% with DeepSeek V4 Flash and 68.53\% with GLM-5.3. Later submissions add 6.71 and 6.03 percentage points, respectively. Among executions that actually enter a retry, 327/1,321 (24.75\%) with DeepSeek V4 Flash and 294/1,064 (27.63\%) with GLM-5.3 eventually pass. Thus, feedback and further attempts recover some failed solutions, but most retried executions still end without a pass. These recovery rates apply only to executions that retry; they do not include every initial failure. The practical implication is that a three-submission allowance provides a useful chance to repair errors while leaving reliable initial execution as the main source of task success.

Looking across frameworks, initial execution and recovery reveal different strengths. OpenCode has the highest first-submission acceptance rate with DeepSeek V4 Flash (64.10\%), whereas DeepSeek Harness starts slightly lower (62.87\%) and gains the most from later submissions (11.49 percentage points). The first result describes solving tasks immediately; the second describes additional success after feedback. A final acceptance rate combines these two behaviors and can hide their difference. However, retry gains alone cannot rank the quality of feedback use: frameworks enter retries on different failed tasks and may spend different amounts of work on each attempt. We therefore report initial acceptance and later recovery separately without equating larger recovery gains with greater efficiency.

The failure categories also need to be read together. Prime Agent has relatively few executions with three failed submissions (9.64\% for DeepSeek V4 Flash and 9.13\% for GLM-5.3), but another 28.92\% and 24.62\% end without a pass after fewer submissions. Looking only at three-submission failures would therefore give an incomplete picture of reliability. The shorter runs record neither successful recovery nor use of the full retry allowance. They can arise from early termination or an exhausted work budget, so the available number of submissions should be distinguished from the number the framework actually completes. Keeping these outcomes separate prevents incomplete attempts from being mistaken for successful error correction.

Within-task retries are available under all three persistence conditions, whereas any post-turn learning call occurs after the submission loop ends. Recovery can therefore demonstrate useful correction within a task without showing that a reusable lesson was saved for future tasks. The traces below and in Appendix~\ref{app:trajectory-audit} help explain this distinction by showing what changed between attempts and whether that change addressed the actual failure.

\subsubsection{Trace-level illustrations of recovery and execution quality}
\label{sec:trace-illustrations}

Two matched task comparisons under DeepSeek V4 Flash Reset show why extra attempts may fail to repair a solution. On L1 Stream 1, Turn 3, Hermes and OpenCode receive the same request for a vulnerability reproducer and start from 51 identical workspace files. Hermes sets up the source path inside the delivered script, which passes the normal invocation and receives reward 1.00. OpenCode tests its script only with a temporary \texttt{PYTHONPATH=src} setting; the normal invocation cannot import the local package, and all three submissions fail the task check (reward 0.00). Its local test therefore checks a different execution environment from the one used to evaluate the delivered script. Repeating this check leaves the mismatch unresolved.

On L2 SRA-blocked Turn 15, DeepSeek Harness uses \texttt{os.path.join} to construct a log-file path and receives reward 1.00. Hermes joins the directory and filename as strings. Its tests use a directory ending in a slash, which hides the defect; when the directory is \texttt{logs}, the output becomes \texttt{logstask\_func.log} rather than \texttt{logs/task\_func.log}. All three submissions retain this defect, and the task receives partial reward 0.15. Here, the failed recovery concerns the test case as well as the code: the test does not expose the input on which path construction fails. Both comparisons show why attempts alone are insufficient: recovery also depends on testing the failing input and repairing the relevant code.

These cases illustrate failure mechanisms; they do not estimate how often those mechanisms occur across the benchmark. Three successful retries in Appendix~\ref{app:retry-success-cases} provide complementary examples: reading earlier accepted artifacts, keeping a correct partial code repair, and revising a recommendation after retrieving a missing fact. None invokes a post-turn learner between submissions. These examples show how agents use available experience to repair the current solution. We next examine saved changes and performance on later tasks.

\subsection{Persistent-State and Probe Analysis (RQ4)}
\label{sec:rq4-reuse}

Successful correction within a task does not show whether the agent saves anything useful for later tasks. We therefore compare recorded changes in learning state with performance on final unseen probes in the DeepSeek V4 Flash and GLM-5.3 audits. This analysis separates having an update procedure, observing a saved change, and finding evidence of useful reuse. State snapshots and probe scores provide complementary evidence.

CLIN tests transfer to new tasks and environments, while Voyager evaluates reuse of its skill library in a new Minecraft world \cite{majumder2024clin,wang2023voyager}. Both assess saved experience through later behavior, motivating our use of unseen probes alongside state inspection. Our probes do not isolate individual update effects.

\begin{figure}[!htb]
    \centering
    \includegraphics[width=\linewidth]{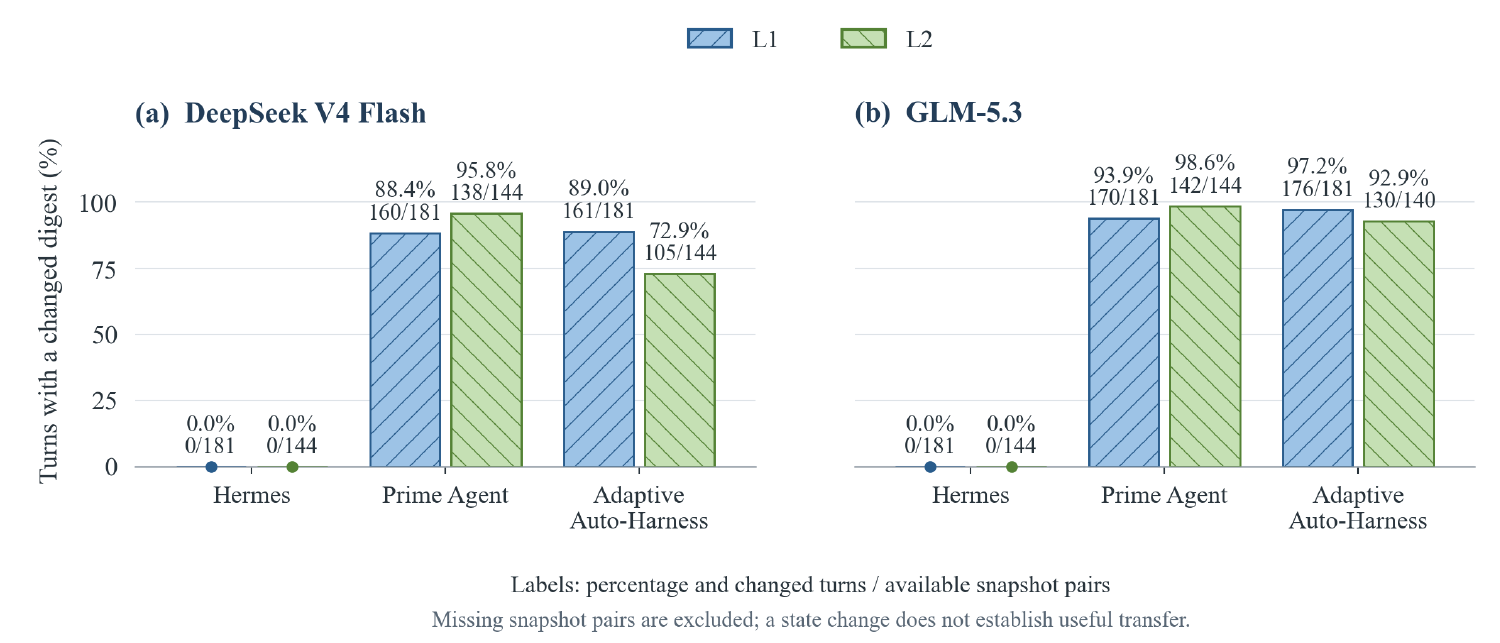}
    \caption{Evolving turns with a changed declared-learning-state digest, grouped by backbone and task level. Labels give changed turns / available before/after snapshot pairs and the corresponding percentage; missing pairs are excluded. A changed digest establishes a recorded state change, not useful transfer.}
    \label{fig:rq4-state-changes}
\end{figure}

We first check whether the declared learning state changes after a turn. \referencehighlight{Figure~\ref{fig:rq4-state-changes}} reports how often the digest, a summary used to detect state changes, differs between available before/after snapshots. Each bar gives changed turns and available snapshot pairs. The denominator matters: for example, GLM-5.3 Adaptive L2 has 140 available pairs rather than 144, so the four missing pairs are excluded from the rate instead of being counted as unchanged.

Prime Agent and Adaptive Auto-Harness frequently change their declared learning state, while Hermes has no changed digest in these records despite having an update interface. OpenCode, DeepSeek Harness, and ZCode have no enabled native post-turn learner in the audited integrations, and their available snapshots remain unchanged. Thus, frameworks evaluated under the same Evolving label do not necessarily perform the same kind of update. Frequent changes confirm that saved state is being revised; they do not show that the revisions are correct or used later. Conversely, an unchanged learning-state digest does not rule out reuse of ordinary task files or context retained outside that declared state.

\begin{figure}[!htb]
    \centering
    \includegraphics[width=0.8\linewidth]{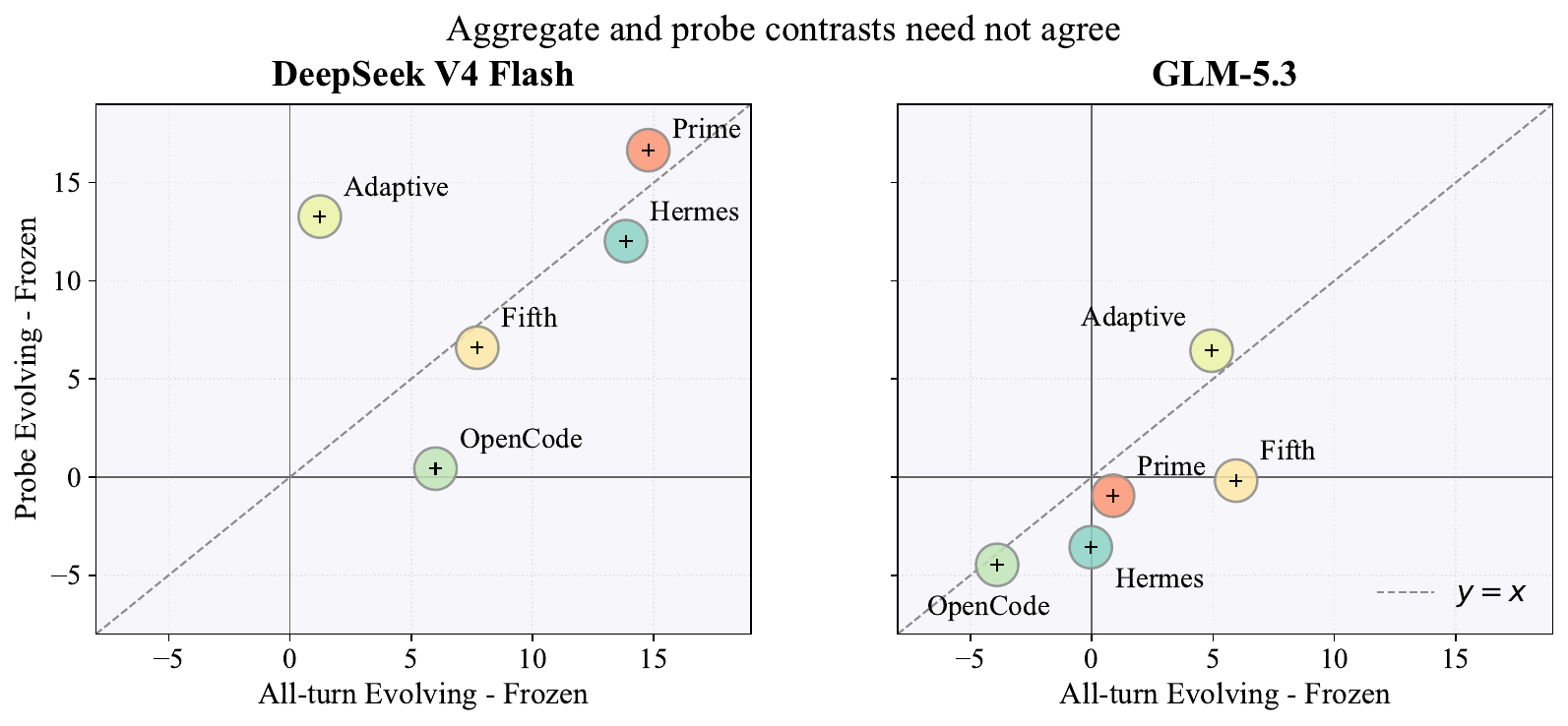}
    \caption{Evolving--Frozen differences on all L2 turns versus the final probe subset for DeepSeek V4 Flash and GLM-5.3. The dashed diagonal marks equal differences. ``Fifth'' denotes DeepSeek Harness for DeepSeek V4 Flash and ZCode for GLM-5.3. Probe differences are computed over 30 final unseen tasks; all-turn differences use 144 turns.}
    \label{fig:probe-vs-allturn}
\end{figure}

\paragraph{The overall score can hide performance on later probes.}
The L2 panel contains 57 learning turns, 57 application turns, and 30 final unseen probes per framework--condition pair. The all-turn mean combines performance while experience is being collected with performance on tasks encountered later. \referencehighlight{Figure~\ref{fig:probe-vs-allturn}} compares the Evolving--Frozen difference over all 144 turns with the difference on the 30 probes. The probe subset lets us check whether the overall advantage also appears on the final unseen tasks.

For DeepSeek V4 Flash Adaptive Auto-Harness, the all-turn difference is $+1.23$ points, while the probe difference is $+13.27$. A small overall difference can therefore hide a larger advantage on the final unseen tasks. The converse also occurs: GLM-5.3 ZCode has a positive all-turn difference ($+5.95$) but almost no probe difference ($-0.18$). GLM-5.3 Prime Agent likewise has a positive all-turn difference ($+0.88$) and a negative probe difference ($-0.95$). These small differences should not be treated as reliable gains or losses without repeated runs, but they show why the sign of the overall mean is insufficient to claim better later performance. Full probe scores are reported in Appendix~\ref{app:trajectory-audit}.

The state audit and probe results thus capture different aspects of experience reuse. Saved changes are observable for some frameworks, but their frequency is not a measure of useful transfer, and a positive all-turn difference need not extend to unseen probes. Even when changes and positive probe differences appear together, the probe tasks differ from earlier tasks and the updated content has not been tested in isolation. Establishing that the saved lesson caused an improvement would require matched probe runs with that lesson retained or removed, under the same execution budget. The current audit reports the two forms of evidence separately and does not include that additional test.

\begin{figure}[!b]
    \centering
    \includegraphics[width=0.8\linewidth]{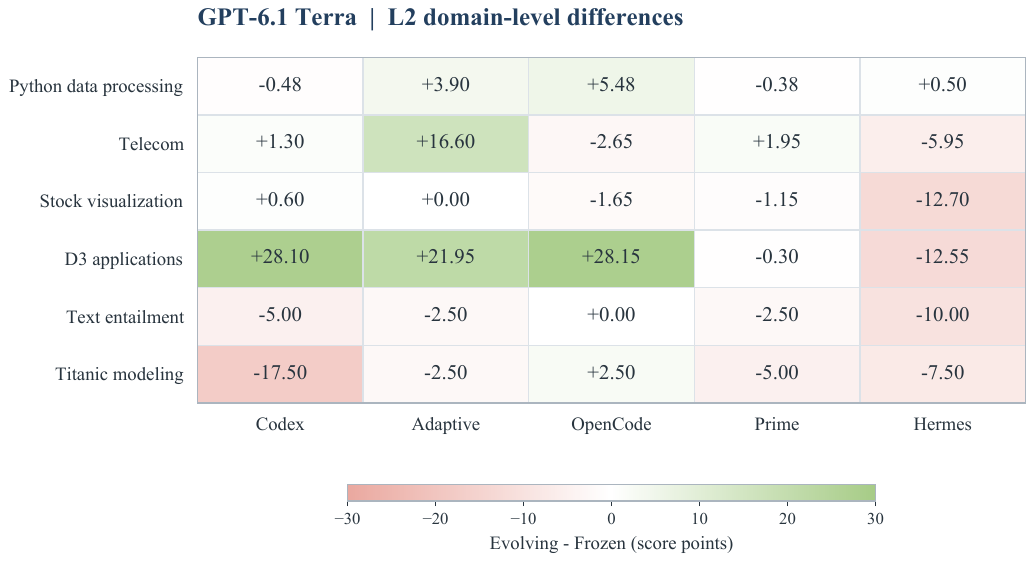}
    \caption{Domain-level Evolving--Frozen score differences on GPT-6.1 Terra L2. Positive values favor Evolving; negative values favor Frozen. Domain sizes are 100, 50, 50, 50, 20, and 20 turns from top to bottom.}
    \label{fig:domain-contrasts}
\end{figure}

\subsection{Domain- and Stream-Level Analysis (RQ5)}
\label{sec:rq5-heterogeneity}

The overall averages leave open how widely the observed gains are shared across tasks. To examine this, we break down GPT-6.1 Terra L2 results by domain and compare independent streams on both L1 and L2. The six L2 domains contain 100 Python data-processing turns, 50 telecom turns, 50 stock-visualization turns, 50 D3-application turns, 20 text-entailment turns, and 20 Titanic-modeling turns. Because the overall score weights turns equally, larger domains contribute more to its value and must be considered when interpreting the aggregate result.

The domain breakdown in \referencehighlight{Figure~\ref{fig:domain-contrasts}} shows that the largest positive differences for Codex, Adaptive, and OpenCode occur in D3 applications ($+28.10$, $+21.95$, and $+28.15$ points). These gains do not extend uniformly to the other domains: Codex is lower by 17.50 points on Titanic modeling, and Adaptive is lower by 2.50 points on both text entailment and Titanic modeling. The overall positive differences therefore combine distinct task outcomes rather than showing the same improvement everywhere. Domain labels help locate these outcomes, but the scores alone do not identify which saved skill or update explains them.

\begin{figure}[!htbp]
    \centering
    \includegraphics[width=\linewidth]{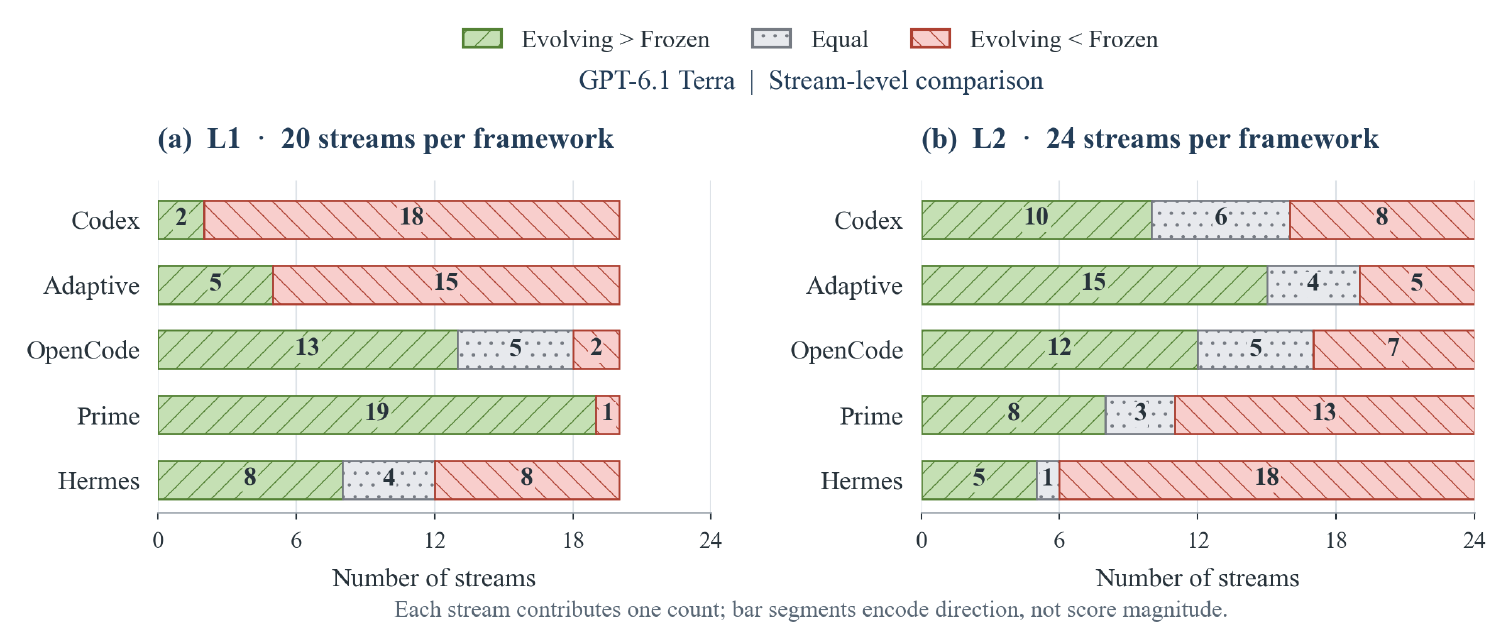}
    \caption{Stream-level Evolving--Frozen directions on GPT-6.1 Terra: positive, equal, and negative counts across 20 L1 and 24 L2 streams. Counts indicate direction, not magnitude.}
    \label{fig:stream-directions}
\end{figure}

To check how much this pattern affects the overall result, we recompute the L2 difference after excluding the 50 D3 turns. Codex changes from $+3.46$ to $-1.68$ points, Adaptive from $+7.65$ to $+4.67$, and OpenCode from $+6.17$ to $+1.59$. Codex's positive overall difference thus depends on including D3 in this task mixture, whereas Adaptive and OpenCode remain positive on the remaining tasks. This distinction separates a result driven by one domain from a result that extends beyond it. Removing a domain changes the evaluation target; this check measures sensitivity to task composition and is not a test of statistical significance.

\paragraph{Stream counts show how widely a gain is shared.}
\referencehighlight{Figure~\ref{fig:stream-directions}} counts streams in which Evolving is higher than, equal to, or lower than Frozen. Adaptive is higher in 15 of 24 L2 streams, giving evidence that its positive mean is accompanied by gains across a majority of the evaluated streams. Codex, by contrast, has a positive mean but is higher in only 10 streams, equal in six, and lower in eight. Its aggregate gain therefore should not be described as improvement on most streams. Each stream contributes one count regardless of its size or score difference, while the overall mean weights turns and preserves the size of each difference. The mean measures the total score advantage; stream counts show how widely that advantage is shared across the evaluated tasks.

The stream results also reinforce the need to separate task levels. Prime is higher in 19 of 20 L1 streams, consistent with its $+22.95$ aggregate difference, but is higher in only eight L2 streams and lower in 13. Its L1 advantage is therefore much more widely shared than its L2 results. This contrast shows why the scope of an improvement must be stated alongside its size: a positive mean may describe large gains on a few tasks, gains across many streams, or both. Domain differences and stream counts show whether gains are limited to selected tasks or shared more widely.

\section{Discussion}
\label{sec:discussion}

\subsection{Self-evolution should be viewed as a continuous process}

The results suggest that self-evolution involves several related but distinct processes, rather than a single update operation. An agent first needs to solve the current task, then identify useful experience from the interaction, and finally apply that experience to later tasks. Within-task retries provide an opportunity to correct the current solution using immediate feedback, while post-turn updates aim to preserve experience beyond the current task. However, neither process alone demonstrates that the agent has improved its future behavior. Later tasks must test this improvement.

The retry results illustrate the difference between local correction and long-term learning. Across DeepSeek V4 Flash and GLM-5.3, 24.75\% and 27.63\% of executions that entered a retry eventually passed. Although later submissions increased the overall acceptance rates, most executions that required retries still failed to pass. This suggests that simply allowing more attempts is not enough. Agents also need to identify why an attempt failed and determine which part of the solution should be changed. The trace examples in Section 5.4 further show that repeated attempts may preserve the same error when the agent fails to test the actual execution conditions or identify the relevant defect.

A similar distinction applies to post-turn updates. Changing persistent memory or skills shows that an agent has recorded new information, but does not establish that the information will improve a later decision. The probe results in Section 5.5 provide a further check: an advantage in the overall L2 score does not always appear on the final unseen tasks. These observations suggest that future self-evolving agents should be designed around a complete learning process: \emph{collect experience, identify useful lessons, update persistent knowledge, and verify its use in subsequent tasks}. Each step should have its own evaluation criteria. For example, failure analysis should examine whether the agent identifies the actual cause of an unsuccessful task, memory updates should be checked for correctness and relevance, and later tasks should test whether the saved experience leads to better decisions or execution.

This view also helps clarify the difference between methods that improve the current response and those that support future tasks. Self-Refine focuses on revising the current output through feedback, whereas CLIN studies adaptation and generalization through persistent textual memory \cite{madaan2023selfrefine,majumder2024clin}. A useful direction is to connect these two processes more closely, so that feedback from current execution can be converted into reusable experience rather than being discarded when the task ends.

\subsection{Knowledge evolution and capability evolution require different learning mechanisms}

The differences between L1 and L2 show that self-evolution cannot be measured or improved through a single type of memory. L1 focuses on information that can be expressed in language, such as facts, preferences, constraints, and procedural rules. L2, in contrast, requires agents to improve their ability to complete tasks through execution, including writing code, processing data, using tools, and working with existing artifacts. Although both levels involve learning from previous experience, the information that needs to be retained and the way it should be reused are different.

The experimental rankings support this distinction. Under DeepSeek V4 Flash Evolving, Adaptive Auto-Harness achieves the highest L1 score (92.15), while DeepSeek Harness leads L2 (80.63). A similar difference appears under GPT-6.1 Terra: Prime Agent reaches 85.91 on L1 but only 37.15 on L2. Moreover, the Evolving--Frozen differences can change direction across the two levels. For example, Prime Agent shows a negative difference on L1 (-6.74 points) but a positive difference on L2 (+14.77 points) under DeepSeek V4 Flash. These results do not imply that one level is inherently more difficult than the other, since they contain different tasks. Instead, they show that an agent's ability to retain explicit knowledge does not necessarily reflect its ability to develop reusable task-solving skills.

This distinction offers a useful direction for future method design. For knowledge evolution, an agent should maintain clear records of facts, user preferences, rules, and their corrections. Each record should include its source, applicable conditions, and validity, so that outdated information can be revised without affecting unrelated knowledge. For capability evolution, simply storing a description of a successful task may not be sufficient. The agent should preserve the procedure that led to success, including the required inputs, execution steps, tool usage, and verification results. New tasks can test this experience, beyond checking that the original artifact remains available.

More importantly, these two forms of learning should not be completely isolated. Explicit knowledge may provide the conditions under which a skill should be used, while procedural experience can provide evidence for revising an existing rule. A promising direction is therefore to maintain different representations for factual knowledge and task-solving skills, while establishing links between them. This would allow an agent to select not only what it remembers, but also which experience is useful for the current task.

\subsection{The quality of an update matters more than its frequency}

The state audit shows that different agent frameworks behave quite differently after task completion. Prime Agent and Adaptive Auto-Harness frequently modify their declared learning states, whereas Hermes has no changed digest in the available records. OpenCode, DeepSeek Harness, and ZCode do not have an enabled native post-turn learner in the audited configurations. These results show that enabling an Evolving condition does not mean that every framework performs the same update operation. Update frequency alone does not show useful learning.

The comparison between all-turn and probe performance provides further evidence. For DeepSeek V4 Flash Adaptive Auto-Harness on L2, the Evolving--Frozen difference is only +1.23 points over all turns but reaches +13.27 points on the final probes. In contrast, GLM-5.3 Prime Agent shows a positive difference of +0.88 points over all turns but a negative difference of -0.95 points on the probes. These examples indicate that aggregate performance and later task performance can reflect different aspects of adaptation. They also suggest that an agent may benefit from updates on some tasks without achieving consistent gains on new tasks. Since the probe results cover only 30 tasks per framework--condition pair and do not isolate individual updates, these differences provide initial, non-causal evidence.

For future self-evolving agents, this raises a more important question: \emph{what should be written into persistent state?} A useful update mechanism should not simply summarize every completed interaction. It should first distinguish general lessons from task-specific details, identify whether a failure is caused by an incorrect decision, a missing tool, or an execution problem, and retain only information that can help with future tasks. Repeated or verified experience may deserve higher priority than a single observation. Meanwhile, outdated rules, duplicated entries, and lessons that conflict with newer evidence should be revised rather than accumulated indefinitely.

The next step is to verify whether the saved experience is actually used. LongMemEval separates memory indexing, retrieval, and reading, showing that retaining information and using it correctly are different challenges \cite{wu2025longmemeval}. A similar design can be adopted for self-evolving agents: record what was saved, whether it was retrieved, how it affected the agent's decision, and whether the resulting task was successful. For procedural skills, the agent should also test a saved procedure on new inputs before treating it as a reliable capability. These checks distinguish useful learning from merely storing more state.

The current probe analysis does not yet establish which particular update leads to a performance change. A stronger evaluation could remove a newly written memory before a probe, disable its retrieval while keeping the memory stored, or compare execution with and without an earlier task artifact. These comparisons could separate gains from learned state, retained workspace artifacts, and existing task-solving ability.

\subsection{Self-evolution should account for task differences and interference}

An agent operates on a variety of tasks over time, and experience that helps in one setting may not be useful in another. The domain- and stream-level results show that an overall performance gain can be concentrated in a small part of the task distribution. In the GPT-6.1 Terra L2 evaluation, Codex, Adaptive Auto-Harness, and OpenCode obtain their largest positive differences on D3 application tasks (+28.10, +21.95, and +28.15 points, respectively). However, these gains do not extend to all other domains. In particular, Codex has a -17.50-point difference on Titanic modeling. When D3 tasks are excluded, its overall Evolving--Frozen difference changes from +3.46 to -1.68 points. This result shows that an aggregate improvement may depend strongly on which tasks are included.

The stream-level results provide another view of this variation. Adaptive Auto-Harness performs better under Evolving in 15 of 24 L2 streams, suggesting that its positive average is shared across a majority of the evaluated streams. Codex, however, performs better in only 10 streams, is equal in six, and performs worse in eight. Prime Agent also shows a clear difference between the two levels: it performs better in 19 of 20 L1 streams but in only eight of 24 L2 streams, with lower scores in 13. Therefore, the size of an average gain and the number of tasks that benefit from it should be considered together. Large gains on a few tasks need not extend across the stream.

These findings point to the importance of task-aware experience management. A self-evolving agent should not retrieve every saved memory or skill whenever it encounters a new request. Instead, it should consider the current task, its required capabilities, available tools, and the conditions under which previous experience was successful. A procedure developed for data processing, for example, should not be applied to an unrelated task simply because both tasks involve code execution. Recording the scope and conditions of a saved skill can help the agent avoid inappropriate reuse.

Task interference also raises the need for better memory maintenance. When a new task introduces a correction or contradicts an earlier rule, the agent should determine whether the change applies globally or only to the current situation. Similarly, frequently used experience should remain accessible, while outdated or rarely useful entries may need to be revised or removed. Future methods could therefore combine task-aware retrieval, conflict checking, and selective memory updates to reduce interference while preserving useful experience over long periods. The goal should not be to accumulate as much information as possible, but to make the right experience available when it is needed.

\subsection{Implications for future evaluation and limitations}

Taken together, the findings suggest that evaluating self-evolving agents requires more than comparing their final task scores. A useful evaluation should examine whether agents can solve tasks initially, recover from unsuccessful attempts, identify and preserve useful experience, and apply that experience to later tasks. It should also report whether improvements are shared across different domains and streams. These measurements provide a clearer view of where an agent improves and where its learning process still fails.

Several limitations should be considered when interpreting the current results. First, the evaluated task cohorts differ across backbone configurations. The DeepSeek V4 Flash and GLM-5.3 evaluations do not include the dedicated artifact-dependency, GPU, and cross-benchmark L2 streams, while GPT-6.1 Terra covers a larger set of tasks and streams. In addition, the fifth framework is configuration-specific: DeepSeek Harness, ZCode, or Codex. Therefore, the results do not support controlled comparisons across backbone models or a universal ranking of all frameworks. Second, the selected historical and replacement runs span different runtime revisions, and a uniform model-call or tool budget has not been certified. Repeated-run confidence intervals are also unavailable, so the reported differences should be interpreted as descriptive observations. Third, detailed trace and persistent-state audits are available for DeepSeek V4 Flash and GLM-5.3, but not for GPT-6.1 Terra. Fourth, the domain and stream analyses are descriptive. Excluding a domain changes the task distribution and should not be interpreted as a test of statistical significance. Finally, the current evaluation does not include systematic removal of source-task artifacts or newly written learning states, so it cannot establish the causal contribution of a particular update to later performance.

These limitations suggest several directions for further evaluation. First, future experiments should use matched task cohorts, consistent execution budgets, and repeated runs to improve the reliability of comparisons. Second, the remaining dependency-focused streams should be included across backbone configurations, allowing experience reuse to be examined under more consistent conditions. Third, state-removal and artifact-removal tests should be introduced to determine whether later success actually depends on previously acquired experience. It would also be useful to extend trace collection to all evaluated frameworks and record memory retrieval, tool calls, model-call and token usage, and execution time. Following the emphasis on cost-aware agent evaluation \cite{kapoor2024agentsmatter}, these measurements could distinguish better experience use from higher resource consumption.

Overall, the results suggest that the next stage of self-evolving agent research should move beyond simply enabling persistent memory or automatic updates. More attention should be given to how agents identify useful experience, represent it in a form that can be reused, select it for the right task, and verify whether it improves later behavior. Developing this complete process is important for building agents that do not merely accumulate experience, but can turn that experience into reliable improvements over long periods of use.

\section{Conclusion}
In this work, we introduce \textbf{EvoLink-Bench}, a benchmark for evaluating self-evolving agents in persistent, interdependent task streams. Its two evaluation levels capture knowledge and capability evolution, while its protocol distinguishes within-task recovery, persistent-state retention, and post-turn updates. Across 19,635 adjudicated turn slots, we find that agent performance varies across evolution levels, persistence conditions, backbone configurations, and task compositions.
Our results show that retries provide limited recovery, persistent-state updates do not necessarily improve subsequent performance, and the benefits of self-evolution vary across domains and task streams. These findings highlight the need to evaluate not only whether agents accumulate experience, but also whether they can effectively reuse it in future tasks. Further causal analysis requires matched-budget repeated runs and controlled removal of previously acquired states or artifacts.

\IfFileExists{references.bib}{%
    {\small
    \numberedauthorbibliography
    \bibliographystyle{plainnat}
    \bibliography{references}
    }%
}{}

\appendix
\section{Implementation and Evaluation Details}
\label{app:experimental-details}

\subsection{Agent State and Update Boundaries}

The benchmark distinguishes task-produced workspace changes from revisions to an agent's declared learning state. Workspace changes occur during ordinary execution and remain available whenever the persistence condition permits. Learning-state changes are restricted to the post-turn update procedure. \referencehighlight{Table~\ref{tab:persistence-conditions}} summarizes this policy for both levels. OpenCode, DeepSeek Harness, and ZCode retain permitted state but have no native learner here.

\begin{table}[!htbp]
    \centering
    \caption{L1 and L2 persistence policies. Learning state includes declared memory, skills, and harness artifacts.}
    \label{tab:persistence-conditions}
    \small
    \setlength{\tabcolsep}{7pt}
    \renewcommand{\arraystretch}{1.25}
    \begin{tabular}{llll}
        \toprule
        \rowcolor{tableheader}
        \textbf{Condition} & \textbf{Workspace} & \textbf{Agent state} & \textbf{Explicit update} \\
        \midrule[\heavyrulewidth]
        Reset    & Reinitialized & Reinitialized & Disabled \\
        Frozen   & Preserved     & Preserved     & Disabled \\
        Evolving & Preserved     & Preserved     & Enabled when implemented \\
        \bottomrule
    \end{tabular}
\end{table}

Hermes reviews completed public trajectories to revise general memory, user memory, and a reusable skill directory. Prime Agent clears the prior conversation and transient execution state at each turn and invokes its refinement interface to revise the persistent harness after scoring. Adaptive Auto-Harness enables prompt, skill, and memory evolution while leaving tool-registry evolution disabled in the reference configuration.

\subsection{Resource Limits}

Public L1 runtime metadata declares ceilings of 12 model calls, 16 tool calls, and 600 seconds for task execution, with a separate turn-boundary learning policy. L2 uses task-specific CPU profiles, with execution ceilings from 600 to 3,600 seconds and tighter model/tool caps where declared. Adapter limits and actual recorded usage must be distinguished from these declarations; the selected historical and replacement runs do not form a certified uniform-budget cohort.

Explicit task failures and deadlines affect only the corresponding turn. Earlier committed rewards and later actually executed turns keep their recorded values. The mix of runtime revisions, incomplete failed-request usage, and differing adapter counters prevents a certified equal-budget or decoding-speed comparison.

\subsection{Task Scoring and Reporting}
\label{sec:experimental-metrics}

A turn's terminal reward $r_{s,t}\in[0,1]$ is distinct from its binary acceptance indicator $b_{s,t}$. The runtime uses the final scored submission rather than selecting the best score over retries. L1 applies package-owned executable checks, including preference checks where specified. L2 uses native verification together with the prescribed task rubric. For MLE tasks declaring continuous native scoring, the judge checks submission validity and a valid result receives the upstream normalized metric, clipped to $[0,1]$. This retains partial credit separately from binary acceptance.

For each configuration and framework, the main tables use a turn-weighted mean,
\begin{equation}
    M_{\ell,c}=\frac{100}{N_{\ell}}\sum_{s\in\mathcal{S}_{\ell}}\sum_{t=1}^{T_s}r_{s,t}^{(c)},
    \qquad N_{\ell}=\sum_{s\in\mathcal{S}_{\ell}}T_s,
\end{equation}
where $\mathcal{S}_{\ell}$ is the set of evaluated streams at level $\ell$ for the given configuration and $c$ is the condition. All recorded turns contribute to the mean. For DeepSeek V4 Flash and GLM-5.3, the L2 score combines 135 primary-panel turns and nine CPU AutoResearch turns. The panel-specific and probe summaries retain their respective turn selections.

In the DeepSeek V4 Flash and GLM-5.3 audit, a recorded execution exception or reached task deadline contributes zero for that turn, while already committed grades and normally returned fractional rewards are preserved. All planned turn slots in these two configurations have an observed grade or an explicit failure decision. Task reward, binary acceptance, and successful learner commitment are separate fields: for example, a valid task grade remains when a post-turn proposal is handled as rejected, even if the raw acceptance flag is false. Saved reports and failure decisions, not optimistic retry selection, define the current values. Appendix~\ref{app:trajectory-audit} gives paired task-level comparisons and learning-state evidence. Explicit failures include execution errors, timeouts, and state-contract violations, not just reasoning errors.

\section{Supplementary Quantitative Analyses}
\label{app:supplementary-quantitative}

This appendix collects expanded condition contrasts, selected prefix checkpoints, and the detailed domain/stream counts underlying RQ2 and RQ5. All values are computed from the saved result summaries described in the main text; they are descriptive and do not constitute repeated-run uncertainty estimates.

\subsection{Condition contrasts across all configurations}

\referencehighlight{Table~\ref{tab:all-condition-contrasts}} reports both the Frozen--Reset persistence contrast and the Evolving--Frozen condition contrast for every framework. The fifth framework is configuration-specific. Contrasts use unrounded means where available.

\begin{table}[!htbp]
    \centering
    \caption{Condition contrasts across the three backbone configurations. $\Delta_{\mathrm{persist}}=M_F-M_R$ and $\Delta_{E-F}=M_E-M_F$. The fifth framework differs across configurations (DeepSeek Harness, ZCode, and Codex).}
    \label{tab:all-condition-contrasts}
    \small
    \setlength{\tabcolsep}{7pt}
    \renewcommand{\arraystretch}{1.25}
    \begin{tabular}{clrrrr}
        \toprule
        \rowcolor{tableheader}
        \textbf{Backbone} & \textbf{Framework} & \textbf{$\Delta_{\rm persist}^{L1}$} & \textbf{$\Delta_{E-F}^{L1}$} & \textbf{$\Delta_{\rm persist}^{L2}$} & \textbf{$\Delta_{E-F}^{L2}$} \\
        \midrule[\heavyrulewidth]
        \multirow{5}{*}{DeepSeek V4 Flash} & Hermes & $+2.32$ & $+6.35$ & $+3.33$ & $+13.85$ \\
         & Prime Agent & $+9.12$ & $-6.74$ & $-3.74$ & $+14.77$ \\
         & Adaptive Auto-Harness & $+5.08$ & $+15.58$ & $+1.19$ & $+1.23$ \\
         & OpenCode & $+5.41$ & $+2.32$ & $-1.04$ & $+6.00$ \\
         & DeepSeek Harness & $+7.57$ & $+4.97$ & $+8.61$ & $+7.72$ \\
        \midrule
        \multirow{5}{*}{GLM-5.3} & Hermes & $+4.70$ & $-2.21$ & $-0.89$ & $-0.04$ \\
         & Prime Agent & $+8.34$ & $-7.90$ & $-1.87$ & $+0.88$ \\
         & Adaptive Auto-Harness & $+5.47$ & $+1.33$ & $+0.74$ & $+4.94$ \\
         & OpenCode & $+0.39$ & $+3.48$ & $+0.54$ & $-3.90$ \\
         & ZCode & $+11.60$ & $+0.22$ & $-1.86$ & $+5.95$ \\
        \midrule
        \multirow{5}{*}{GPT-6.1 Terra} & Hermes & $+5.18$ & $+0.35$ & $+8.46$ & $-6.41$ \\
         & Prime Agent & $+10.59$ & $+22.95$ & $+2.14$ & $-0.56$ \\
         & Adaptive Auto-Harness & $+13.17$ & $-3.04$ & $+0.09$ & $+7.65$ \\
         & OpenCode & $+36.37$ & $+5.31$ & $+0.07$ & $+6.17$ \\
         & Codex & $+9.79$ & $-3.33$ & $-0.38$ & $+3.46$ \\
        \bottomrule
    \end{tabular}
\end{table}

\subsection{Selected temporal prefix contrasts}

\referencehighlight{Table~\ref{tab:prefix-checkpoints}} lists selected Evolving--Frozen differences at 25\%, 50\%, and 100\% of stream progress. These examples were selected to illustrate distinct temporal patterns, including a sign reversal and a late-emerging positive difference. A prefix contains the tasks encountered up to that point; it is not a repeated evaluation of a fixed task set.

\begin{table}[!htbp]
    \centering
    \caption{Evolving--Frozen differences (score points) at 25\%, 50\%, and 100\% of selected streams.}
    \label{tab:prefix-checkpoints}
    \small
    \setlength{\tabcolsep}{7pt}
    \renewcommand{\arraystretch}{1.25}
    \begin{tabular}{llcrrr}
        \toprule
        \rowcolor{tableheader}
        \textbf{Backbone} & \textbf{Framework} & \textbf{Level} & \textbf{25\%} & \textbf{50\%} & \textbf{100\%} \\
        \midrule[\heavyrulewidth]
        DeepSeek V4 Flash & Adaptive & L1 & $+15.61$ & $+10.34$ & $+15.58$ \\
        DeepSeek V4 Flash & Prime & L1 & $-5.37$ & $-8.09$ & $-6.74$ \\
        DeepSeek V4 Flash & Adaptive & L2 & $+4.30$ & $-4.68$ & $+1.23$ \\
        DeepSeek V4 Flash & Prime & L2 & $+4.62$ & $+13.98$ & $+14.77$ \\
        GLM-5.3 & OpenCode & L2 & $+3.25$ & $-0.91$ & $-3.90$ \\
        GLM-5.3 & ZCode & L2 & $-3.12$ & $+0.15$ & $+5.95$ \\
        \bottomrule
    \end{tabular}
\end{table}

\subsection{Detailed domain-level differences on GPT-6.1 Terra L2}

\begin{figure}[!htbp]
    \centering
    \includegraphics[width=\linewidth]{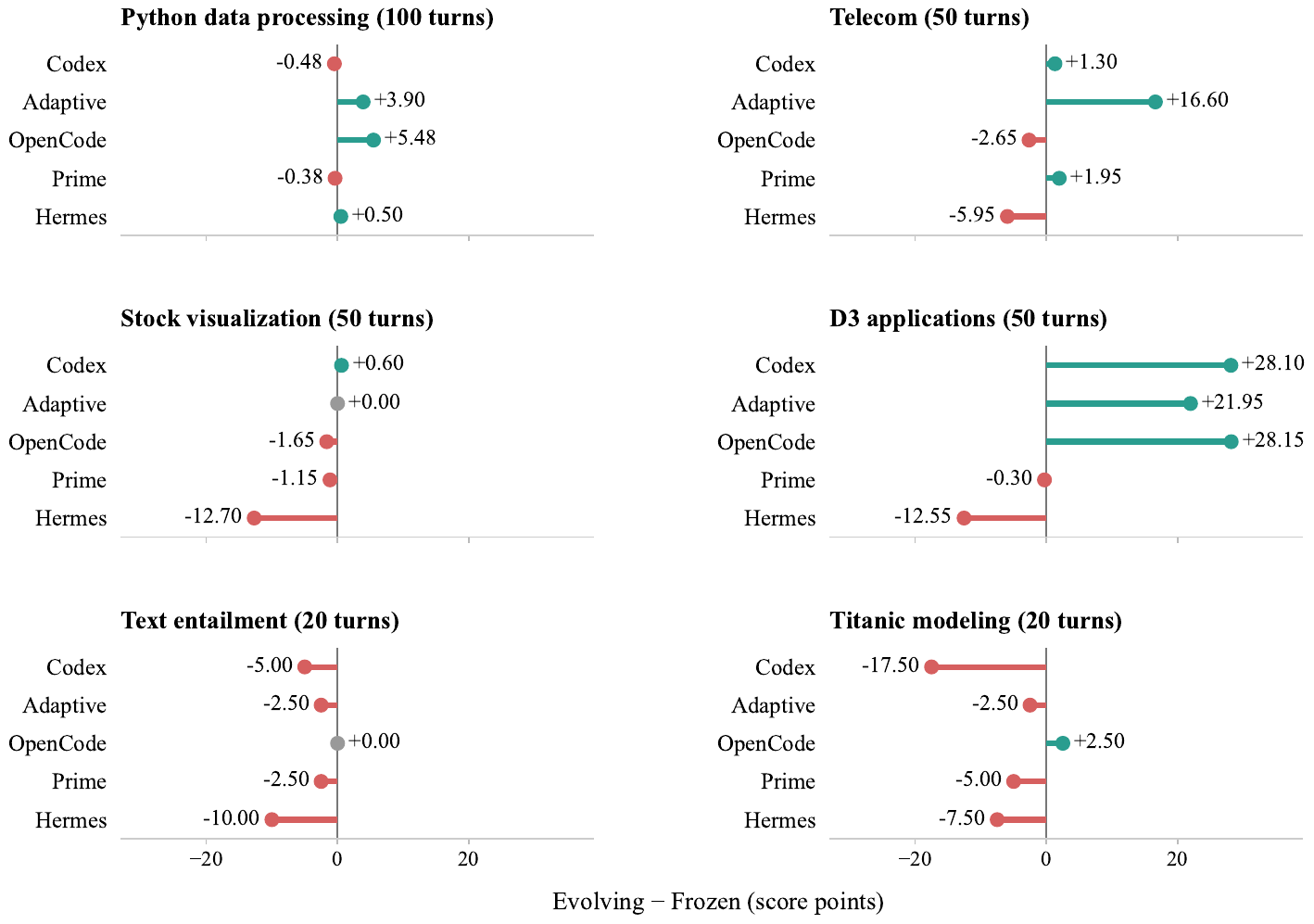}
    \caption{GPT-6.1 Terra L2 domain-level Evolving--Frozen differences. Each segment connects zero to a framework's signed score difference; all six panels share the same horizontal scale.}
    \label{fig:domain-delta-detail}
\end{figure}

\referencehighlight{Figure~\ref{fig:domain-delta-detail}} expands the domain-level comparison in \referencehighlight{Figure~\ref{fig:domain-contrasts}} into six profiles on a common scale. Rightward marks indicate positive Evolving--Frozen differences and leftward marks indicate negative differences; labels retain the exact reported values. Panel titles give the number of turns in each domain.

\subsection{Stream-level direction counts on GPT-6.1 Terra}

\begin{figure}[!htbp]
    \centering
    \includegraphics[width=\linewidth]{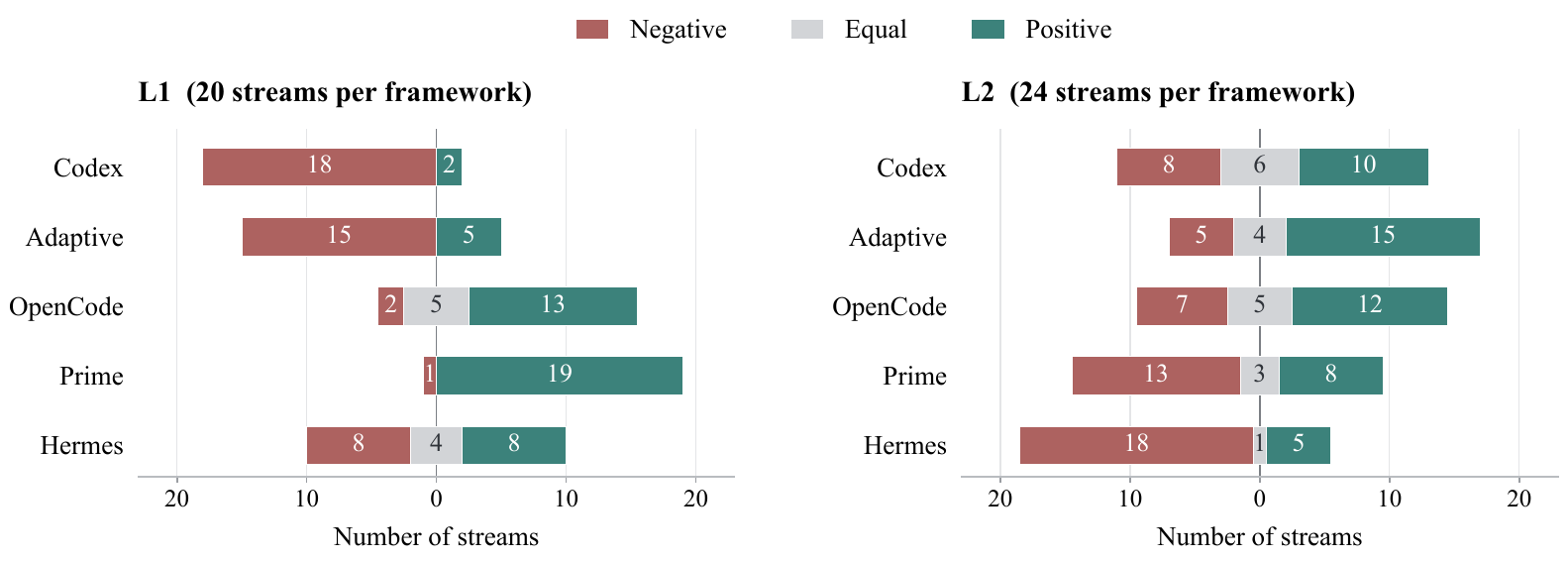}
    \caption{Stream-level Evolving--Frozen directions on GPT-6.1 Terra. Segment widths and labels give the number of negative (left), equal (center), and positive (right) streams. An absent segment indicates zero streams. Counts indicate direction, not the size of the score difference.}
    \label{fig:stream-direction-detail}
\end{figure}

\referencehighlight{Figure~\ref{fig:stream-direction-detail}} shows the positive, equal, and negative stream counts underlying \referencehighlight{Figure~\ref{fig:stream-directions}} as diverging bars on a common count scale. The unit is an independent stream, and the sign is based on its mean Evolving--Frozen difference. Equal counts are centered at zero; negative and positive counts extend to the left and right, respectively.

\section{Trace-Level Audits of Execution and Persistent Updates}
\label{app:trajectory-audit}

\subsection{Current Scope and Source Selection}

This audit covers 300 cells and 4,875 turn slots for each of DeepSeek V4 Flash and GLM-5.3. DeepSeek V4 Flash retains 211 unaffected cells and substitutes the fixed 89-cell repair scope with complete attempts. GLM-5.3 retains its committed prefixes and fills 92 successor slots, with 21 total exception/timeout decisions across the complete selection. Normal rewards and acceptance flags remain bound to their source records. Task rewards remain separate from learning-completion flags; an unsuccessful update does not automatically erase a valid task evaluation.

\begin{table}[!htbp]
    \centering
    \caption{Current evidence coverage. Counts include submissions on graded turns, excluding incomplete and superseded attempts. Fractional rewards are valid scores between 0 and 1.}
    \label{tab:trajectory-audit-coverage}
    \small
    \setlength{\tabcolsep}{7pt}
    \renewcommand{\arraystretch}{1.25}
    \begin{tabular}{lrr}
        \toprule
        \rowcolor{tableheader}
        \multicolumn{1}{c}{\textbf{Evidence}} & \textbf{DeepSeek V4 Flash} & \textbf{GLM-5.3} \\
        \midrule[\heavyrulewidth]
        Selected stream cells & 300 & 300 \\
        Scored turn slots & 4,875 & 4,875 \\
        Normally graded terminal records & 4,875 & 4,854 \\
        Explicit exception/timeout zeros & 0 & 21 \\
        Submission records attached to graded turns & 7,197 & 6,713 \\
        Normal fractional rewards retained & 747 & 703 \\
        \bottomrule
    \end{tabular}
\end{table}

The DeepSeek V4 Flash failure count in this table refers to explicit whole-turn adjudication decisions, not a claim that every historical interaction succeeded. The audit does not fill unknown usage or missing state observations with zero. Current machine-readable sources are the main-table release under \path{runs/deepseek-glm-main-tables-20261105/} and the GLM-5.3 turn-scoring release under \path{runs/glm53-final-turn-scoring-20261105/}. The offline inventory records report hashes, turn identities, and calculations; this audit replays no tasks or paid interactions.

\subsection{Case 1: Deliverable Verification Under the Normal Entry Point}
\label{app:delivery-case}

DeepSeek V4 Flash Hermes and OpenCode are compared on L1 Stream 1, Turn 3, under Reset. Both records belong to the same batch and have the same initial request and 51 initial workspace file hashes. The request is to produce a standalone vulnerability-reproduction script. \referencehighlight{Table~\ref{tab:paired-capability-cases}} preserves the normal task rewards and recorded submission counts for both this pair and the logging pair below.

\begin{table}[!htbp]
    \centering
    \caption{Same-task baseline comparisons with the DeepSeek V4 Flash backbone and Reset condition. Scores are the original 0--1 task rewards, including the retained 0.15 partial reward. Each pair fixes the task request and run batch; no cross-task learning claim or new evaluation score is inferred.}
    \label{tab:paired-capability-cases}
    \small
    \setlength{\tabcolsep}{7pt}
    \renewcommand{\arraystretch}{1.25}
    \begin{tabular}{llrr}
        \toprule
        \rowcolor{tableheader}
        \multicolumn{1}{c}{\textbf{Task}} & \textbf{Framework} & \textbf{Submissions} & \textbf{Reward} \\
        \midrule[\heavyrulewidth]
        L1 stream 1, turn 3 & Hermes & 1 & 1.00 \\
         & OpenCode & 3 & 0.00 \\
        \addlinespace[2pt]
        L2 SRA-blocked, turn 15 & DeepSeek Harness & 1 & 1.00 \\
         & Hermes & 3 & 0.15 \\
        \bottomrule
    \end{tabular}
\end{table}

Hermes derives the repository location from \texttt{\_\_file\_\_} and adds the source directory to \texttt{sys.path} inside the delivered script. Its normal invocation, \texttt{python3 scripts/reproduce\_cve.py}, passes the task check. OpenCode imports \texttt{get\_netrc\_auth} from \texttt{requests.utils}, but its saved script does not configure the local source path. Both of its own execution commands set \texttt{PYTHONPATH=src}; the independent plain invocation instead reports \texttt{ModuleNotFoundError}. The third submission only inspects the script, leaving the setup error unresolved and the final reward at 0.0.

The verification error is an environment mismatch between local testing and normal script execution.

\subsection{Case 2: Path Handling and a Defect Left Unchanged Across Retries}
\label{app:logging-case}

The L2 pair uses SRA-blocked \texttt{s01}, Turn 15, under Reset, with the same public request, runtime-profile digest, and effective work limits. The function must count a requested element and write \texttt{task\_func.log} inside the directory supplied as \texttt{log\_path}. DeepSeek Harness correctly builds this path with \texttt{os.path.join(log\_path, "task\_func.log")}.

All three saved Hermes submissions use direct string concatenation, equivalent to \texttt{log\_path + "task\_func.log"}. With \texttt{log\_path="logs"}, this gives \texttt{logstask\_func.log}, while the required destination is \path{logs/task_func.log}. The agent's own tests use \texttt{./}, where the separator is already present. Its later edits focus on closing or clearing logging handlers and leave the path expression unchanged. Hermes retains partial reward 0.15; DeepSeek Harness scores 1.0.

The existing offline artifact audit checks the saved DeepSeek Harness version and all three saved Hermes versions with and without a trailing slash. It confirms the output-location difference while preserving the returned count. These checks use temporary logging paths and do not create a new benchmark score. The evidence supports a concrete path-handling and revision-strategy contrast; it does not establish that either framework is always better on unrelated tasks.

\subsection{Recorded Learning-State Changes}

The main-text \referencehighlight{Figure~\ref{fig:rq4-state-changes}} reports the changed-digest counts and rates for the available Evolving snapshots. The native-execution baselines have no enabled post-turn learner in these integrations, and their available declared-learning-state snapshots remain unchanged. Hermes also has no changed digest in the selected DeepSeek V4 Flash and GLM-5.3 records, despite nonzero score differences between conditions. These observations reinforce that a condition label or score difference cannot be used as a substitute for direct update evidence.

The Prime Agent GLM-5.3 records contain 172 state-changing updates whose reported learner-token count is zero. This counter does not establish zero-cost refinement. For both Adaptive and Prime, a digest change confirms a saved update; later tasks must test its accuracy, retrieval, and usefulness.

\subsection{Submission Counts and Persisted Failure Lessons}
\label{app:retry-outcomes}

The retry census covers the 600 selected stream cells and 9,750 turn executions reported for DeepSeek V4 Flash and GLM-5.3. It reads 13,910 submission records attached to 9,729 normally graded turns; the remaining 21 GLM-5.3 slots are explicit exception/timeout zeros. No attempt count is invented for these exception slots. Each normal turn has one to three recorded submissions. The unit is a task execution under one backbone, framework, and condition, rather than a distinct source problem. \referencehighlight{Table~\ref{tab:retry-outcomes-overall}} gives the full count partition.

\begin{table}[!htbp]
    \centering
    \caption{Submission-level task acceptance and non-acceptance, with 4,875 executions per backbone. Percentages use all selected executions. Valid partial rewards are retained for unaccepted submissions.}
    \label{tab:retry-outcomes-overall}
    \small
    \setlength{\tabcolsep}{7pt}
    \renewcommand{\arraystretch}{1.25}
    \begin{tabular}{lrrrr}
        \toprule
        \rowcolor{tableheader}
         & \multicolumn{2}{c}{\textbf{DeepSeek V4 Flash}} & \multicolumn{2}{c}{\textbf{GLM-5.3}} \\
        \rowcolor{tableheader}
        \multicolumn{1}{c}{\multirow{-2}{*}{\textbf{Outcome}}} & \textbf{Count} & \textbf{\%} & \textbf{Count} & \textbf{\%} \\
        \midrule[\heavyrulewidth]
        Pass on submission 1 & 2,910 & 59.69 & 3,341 & 68.53 \\
        Pass on submission 2 & 238 & 4.88 & 215 & 4.41 \\
        Pass on submission 3 & 89 & 1.83 & 79 & 1.62 \\
        No pass after 3 submissions & 912 & 18.71 & 716 & 14.69 \\
        No pass with fewer than 3 & 726 & 14.89 & 503 & 10.32 \\
        Explicit exception/timeout zero & 0 & 0.00 & 21 & 0.43 \\
        \bottomrule
    \end{tabular}
\end{table}

\paragraph{Feedback within a task is not yet a learned rule.}
After a normally evaluated submission, the runner saves the response, tool calls, reward, acceptance flag, and public failure feedback. L1 records these in \texttt{turns[].interactions[]} in the summary or retained completed-turn prefix; L2 records them in \path{trace.jsonl}. The feedback enters the follow-up request or dialogue history for the next submission. Task files can also remain available for a local correction. This preserves the evidence needed to retry the current task, but does not by itself create a reusable rule for future tasks.

\paragraph{The explicit learning call occurs after the turn.}
The runner invokes the shared learning barrier once the task's submission loop has ended, passing the public trajectory and feedback from that turn. It does not invoke this cross-task learner after every failed submission. Reset and Frozen disable this update path. Evolving enables supported learners, but an enabled call can still fail, be rejected, or leave the state unchanged.

Adaptive Auto-Harness can write \path{memory/memories.jsonl}, \path{prompts/system.md}, and \path{skills/<name>/SKILL.md}; it also writes an audit history. Prime Agent stores harness updates in \path{prime-home/harness/harness_state.json} and the update ledger \path{prime-home/harness/refinements.jsonl}. The checkpoint stores committed learning-state bytes in \path{baseline_state.zip}, separately from the task workspace in \path{workspace.zip}. Failed drafts may remain under \path{failed_evolution/}, but official state is rolled back; drafts are not committed updates.

\begin{table}[!htbp]
    \centering
    \caption{Recorded persistent learning-file changes after Evolving turns containing at least one failed submission. Denominators include normally recorded Evolving turns with a failed submission, including turns later repaired by retry. Adaptive counts changes to memory, prompt, skill, or tool paths and excludes audit-history-only changes; Prime counts declared harness-state changes. Missing records are excluded; file changes alone do not prove learning.}
    \label{tab:failure-learning-files}
    \small
    \setlength{\tabcolsep}{7pt}
    \renewcommand{\arraystretch}{1.25}
    \begin{tabular}{lrr}
        \toprule
        \rowcolor{tableheader}
        \multicolumn{1}{c}{\textbf{Framework}} & \textbf{DeepSeek V4 Flash} & \textbf{GLM-5.3} \\
        \midrule[\heavyrulewidth]
        Hermes & 0/125 & 0/93 \\
        Prime Agent & 119/123 & 130/133 \\
        Adaptive Auto-Harness & 73/110 & 86/93 \\
        OpenCode & 0/100 & 0/109 \\
        DeepSeek Harness / ZCode & 0/95 & 0/64 \\
        \bottomrule
    \end{tabular}
\end{table}

\referencehighlight{Table~\ref{tab:failure-learning-files}} shows that failure feedback is not uniformly turned into persistent updates. Adaptive changes memory, prompt, or skill-related paths after 73/110 such turns with DeepSeek V4 Flash and 86/93 with GLM-5.3. Including audit-history-only changes would instead give 76/110 and 87/93. Prime records harness-state changes after 119/123 and 130/133. Hermes has no changed committed learning-state digest in these records; several learning attempts do not complete and are rolled back. OpenCode, DeepSeek Harness, and ZCode have no enabled native post-turn learner in these integrations. None of these counts proves that the next task uses an update or benefits from it.

\paragraph{Two saved examples show what is actually written.}
In DeepSeek V4 Flash Adaptive's L2 Evolving SRA-interleaved stream (\texttt{s02}), Turn 4 fails all three task submissions and retains reward 0.15. After the turn, the learner writes a memory entry and a \path{skills/code-only-submission/SKILL.md} rule advising raw source-code output without Markdown fences or extra explanation. In DeepSeek V4 Flash Prime's L2 Evolving SRA-blocked stream (\texttt{s01}), Turn 6 also fails all three submissions and retains reward 0.075. Its saved harness update advises plain built-in return types rather than custom compatibility classes, and avoiding irrelevant workspace searches. These are the agents' own summaries, not independent verification that their diagnosis is correct.

For both examples, the state-archive member bytes match the recorded post-update hashes and the same bytes appear in the following turn's pre-learning state. This establishes persistence. It does not establish that the model read, followed, or benefited from the rule. The shareable numerical provenance, including the complete model--level--system--condition breakdown, is provided in \path{sections/current-retry-summary.json}.

\subsection{Cases That Pass After One or Two Retries}
\label{app:retry-success-cases}

What does a successful retry actually reuse? We inspect three normally graded examples from the current Evolving selection, chosen from 72 retry successes screened for Adaptive Auto-Harness and Prime Agent across both backbones. These are illustrative cases, not a representative sample or a baseline ranking. One retry means two submissions in total; two retries mean three. \referencehighlight{Table~\ref{tab:retry-success-cases}} preserves every submission's native reward. Earlier partial rewards are shown as recorded, rather than replaced with zero. The final submission is accepted in all three cases.

\begin{table}[!htbp]
    \centering
    \caption{Selected retry successes. All three use Evolving. Rewards are the saved submission-level values, not new evaluations. R1 reuses earlier successful task artifacts; R2 and R3 show corrections within the current task.}
    \label{tab:retry-success-cases}
    \small
    \setlength{\tabcolsep}{7pt}
    \renewcommand{\arraystretch}{1.25}
    \begin{tabular}{llll}
        \toprule
        \rowcolor{tableheader}
        \textbf{Case} & \textbf{Backbone / framework} & \textbf{Task location} & \textbf{Rewards in order} \\
        \midrule[\heavyrulewidth]
        R1 & DeepSeek V4 Flash / Adaptive & L1 s14, turn 16 & $0 \rightarrow 1$ \\
        R2 & GLM-5.3 / Adaptive & L2 s01, turn 14 & $0.4 \rightarrow 0.275 \rightarrow 1$ \\
        R3 & GLM-5.3 / Prime & L2 s05, turn 15 & $0.275 \rightarrow 0.275 \rightarrow 1$ \\
        \bottomrule
    \end{tabular}
\end{table}

\subsubsection{R1: Reusing Earlier Successful Reports to Finish a Summary}

The task asks DeepSeek V4 Flash Adaptive to write \path{rca_outputs/executive_summary.md}, combining the earlier reports in a recall-analysis workspace and including four required YAML fields.

\paragraph{Saved trajectory, followed by its interpretation.}
On submission 1, the agent lists the output directory, reads prior JSON artifacts and supplied source documents, and inspects headings in earlier reports. It returns an empty final response and the requested summary file is absent from the saved post-submission file inventory. The native task reward is 0. The record establishes an incomplete deliverable; it does not establish why the agent stopped before writing it.

The next public request says the summary is still unfinished and asks the agent to write the file, consolidate the prior rounds, and verify its existence. On submission 2, the saved tool calls read \path{nimbus_recall_metadata.json}, \path{capa_comparison.json}, the revised metadata, and other retained outputs. A separate call reads the frontmatter of \path{nimbus_5why.md} and \path{ivenix_tech_summary_v1.md}. The agent then writes the summary, checks its required YAML fields, checks source citations, and repairs citation paths. The saved summary passes with reward 1.

This case provides a concrete link to earlier successes. The metadata from Turn 1, report from Turn 3, technical summary from Turn 8, comparison from Turn 10, and revised metadata from Turn 14 each belong to a task accepted with reward 1. Their saved post-task file hashes match the corresponding files available before the successful retry. The retry's tool results show those artifacts being read, rather than merely present in the workspace.

\paragraph{Why the retry helps.}
The agent already had much of the information on its first attempt. What changes is that it turns the retained work into the required deliverable and checks it. Earlier successful files provide reusable facts, structure, and formatting examples; the failure feedback redirects attention to the missing output. This is observed reuse of successful task artifacts across turns. It does not show that a learned memory rule caused the recovery: artifact retention can also support reuse without a post-turn learner, and no artifact-removal comparison is available.

\subsubsection{R2: Keeping a Correct Fix and Repairing the Remaining Requirement}

GLM-5.3 Adaptive's SRA-blocked \texttt{s01}, Turn 14 (\texttt{bigcodebench\_00956}), asks for a string function. It must remove punctuation, apply seeded random casing, and map spaces to one underscore, tabs to two, and newlines to three.

\paragraph{Saved trajectory, followed by its interpretation.}
Submission 1 inserts underscores for whitespace and then removes punctuation. Because an underscore is itself punctuation, the second operation deletes the separators that the first operation just created. The native reward is 0.4. The first retry moves punctuation removal before whitespace conversion, preserving the inserted separators. However, it uses a single regular-expression replacement for all whitespace, losing the distinction between spaces, tabs, and newlines. Despite claiming success, it fails with reward 0.275.

The second retry receives an explicit public reminder of the three required replacement lengths. It keeps the corrected operation order and replaces newline, tab, and space separately. The native task now passes with reward 1. A small offline check of the three saved code blocks makes the difference visible: for input \verb|a b\tc\nd!|, where \verb|\t| and \verb|\n| denote tab and newline, the lowercase outputs are respectively \verb|abcd|, \verb|a_b_c_d|, and \verb|a_b__c___d|. Lowercasing here only makes the whitespace behavior easier to read. The check leaves the native score unchanged.

\paragraph{Why the retry helps.}
The agent uses a useful part of its previous attempt: punctuation is still removed first. It then changes the part that remains wrong, instead of discarding the whole solution. The detailed follow-up supplies the remaining requirement directly, so this is evidence of using task feedback and retaining a successful local correction, rather than unaided discovery of a new rule. The agent also spends calls browsing task/workspace files and initially tests the wrong expectation that all whitespace becomes one underscore. Those actions do not explain the recovery; the decisive visible change is the final replacement logic. The drop in intermediate reward also shows that a plausible edit and a confident self-report do not guarantee an improvement.

\subsubsection{R3: Correcting a Recommendation by Checking the Missing Fact}

GLM-5.3 Prime's tau2-interleaved \texttt{s05}, Turn 15 (\texttt{banking\_knowledge:task\_001}), is a conversation about personal-card options in a fictional banking knowledge base. The live conversation, rather than the final summary, is graded.

\paragraph{Saved trajectory, followed by its interpretation.}
Submission 1 ends with the customer choosing the Silver card, but receives only 0.275 and is not accepted. Submission 2 again receives 0.275. Its final summary describes the Crypto-Cash Back card as having no annual fee, and the saved client output independently shows the customer choosing that card on precisely that understanding. Customer satisfaction and dialogue completion do not prove a correct recommendation.

On submission 3, a saved knowledge-base tool result explicitly lists an annual fee of \$75 for that card. Further tool outputs retrieve alternative card terms and the \$4.99 monthly premium-membership cost. The final client output records the customer choosing the Gold card and \texttt{ended: true}; the native task passes with reward 1. The agent's final summary also identifies its earlier fee error and says it disclosed the alternative's subscription requirement. This self-description is consistent with the changed lookup and outcome, but is not treated as a verbatim record of everything said to the customer: the normalized trace stores empty arguments for these tool calls.

\paragraph{Why the retry helps.}
Repeating a recommendation was insufficient. The successful attempt retrieves the fact that undermines the previous answer and checks the costs of the alternative. Unlike R2's final follow-up, this retry request does not name the incorrect fee or provide the correct recommendation; it asks the agent to follow the policy and finish the task correctly. The visible evidence supports recovery through additional knowledge-base checking, with the agent explicitly relating the correction to its earlier answer. It does not isolate that fee as the sole reason for the score change: each attempt regenerates the conversation, without single-factor replay. Card facts are supplied fixtures.

\paragraph{What these cases establish about experience.}
R1 shows earlier successful work being read and reused. R2 shows a correct partial repair surviving into the next revision, with specific feedback resolving the remaining mistake. R3 shows a previous answer being reconsidered against newly retrieved evidence. These are different, observable ways to use experience. In none of them is the post-turn learner called between the failed submission and its retry. The cases therefore explain successful recovery without treating it as proof of a newly persisted cross-task rule. The persistence examples in Appendix~\ref{app:retry-outcomes} answer a separate question: what is saved after the entire turn. Public case identities, report/trace hashes, record locations, and rewards are listed in \path{sections/current-retry-case-summary.json}.

\subsection{All-Turn and Probe Sensitivity}

Each L2 baseline--condition pair has 57 learning turns, 57 application turns, and 30 probes. \referencehighlight{Table~\ref{tab:audit-probe-sensitivity}} uses the same saved rewards, including explicit failure zeros when they fall on probes. It is a metric-scope comparison rather than a new experiment or a repeated-run significance test.

\begin{table}[!htbp]
    \centering
    \caption{Current L2 probe sensitivity. Probe scores use 30 turns per condition; all-turn differences use 144. $\Delta$ is Evolving minus Frozen, computed from unrounded values.}
    \label{tab:audit-probe-sensitivity}
    \small
    \setlength{\tabcolsep}{7pt}
    \renewcommand{\arraystretch}{1.25}
    \begin{tabular}{clrrrr}
        \toprule
        \rowcolor{tableheader}
         &  & \multicolumn{2}{c}{\textbf{Probe score}} & \multicolumn{2}{c}{\textbf{$\Delta$}} \\
        \rowcolor{tableheader}
        \multirow{-2}{*}{\textbf{Backbone}} & \multirow{-2}{*}{\textbf{Framework}} & \textbf{Frozen} & \textbf{Evolving} & \textbf{All turns} & \textbf{Probes} \\
        \midrule[\heavyrulewidth]
        \multirow{5}{*}{DeepSeek V4 Flash} & Hermes & 52.48 & 64.50 & $+13.85$ & $+12.02$ \\
         & Prime Agent & 46.08 & 62.73 & $+14.77$ & $+16.65$ \\
         & Adaptive Auto-Harness & 47.25 & 60.52 & $+1.23$ & $+13.27$ \\
         & OpenCode & 56.57 & 57.00 & $+6.00$ & $+0.43$ \\
         & DeepSeek Harness & 76.58 & 83.17 & $+7.72$ & $+6.59$ \\
        \midrule
        \multirow{5}{*}{GLM-5.3} & Hermes & 72.89 & 69.31 & $-0.04$ & $-3.57$ \\
         & Prime Agent & 57.42 & 56.47 & $+0.88$ & $-0.95$ \\
         & Adaptive Auto-Harness & 70.42 & 76.86 & $+4.94$ & $+6.44$ \\
         & OpenCode & 76.43 & 71.96 & $-3.90$ & $-4.47$ \\
         & ZCode & 74.31 & 74.13 & $+5.95$ & $-0.18$ \\
        \bottomrule
    \end{tabular}
\end{table}

The GLM-5.3 Prime and ZCode contrasts have opposite signs in the all-turn and probe views. For DeepSeek V4 Flash Adaptive Auto-Harness, the probe difference is much larger than the all-turn difference. A task aggregate, a held-out-capability score, and a causal update effect are therefore different quantities. The selected subset excludes the dedicated artifact-dependency, GPU, and cross-benchmark panels, and no source-episode deletion tests whether a task becomes harder or incomplete without an antecedent.

\clearpage
\end{document}